\documentclass{article}

\usepackage{tikz}
\usepackage{pgfplots}
\usepackage{PRIMEarxiv}
\usepackage{amsmath,amssymb,amsfonts}
\usepackage[utf8]{inputenc} 
\usepackage[T1]{fontenc}    
\usepackage{hyperref}       
\usepackage{url}            
\usepackage{booktabs}       
\usepackage{amsfonts}       
\usepackage{nicefrac}       
\usepackage{subcaption}
\usepackage{multirow}
\usepackage{color,soul}
\usepackage{microtype}      
\usepackage{lipsum}
\usepackage{fancyhdr}       
\usepackage{graphicx}       
\graphicspath{{media/}}     
\usepackage{float}

\title{sEMG-Driven Physics-Informed Gated Recurrent Networks for Modeling Upper Limb Multi-Joint Movement Dynamics
}

\author{
  Rajnish Kumar\\
  Department of Applied Mechanics \\
  Indian Institute of Technology \\
  New Delhi\\
  \texttt{Rajnish.Kumar@am.iitd.ac.in}\\
  \And
  Anand Gupta\\
  Department of Applied Mechanics \\
  Indian Institute of Technology \\
  New Delhi\\
  \texttt{ama222279@iitd.ac.in}\\
   \And
  Suriya Prakash Muthukrishnan \\
  Department of Physiology \\
  All India Institute of Medical Sciences \\
  New Delhi\\
  \texttt{dr.suriyaprakash@aiims.edu} \\
   \And
  Lalan Kumar \\
  Department of Electrical Engineering \\
  Indian Institute of Technology\\
  New Delhi\\
  \texttt{lalank@ee.iitd.ac.in} \\
   \And
  Sitikantha Roy \\
  Department of Applied Mechanics \\
  Indian Institute of Technology\\
  New Delhi\\
  \texttt{sroy@am.iitd.ac.in}
}

\begin{document}
\maketitle

\begin{abstract}
Exoskeletons and rehabilitation systems have the potential to improve human strength and recovery by using adaptive human-machine interfaces. Achieving precise and responsive control in these systems depends on accurately estimating joint movement dynamics, such as joint angle, velocity, acceleration, external mass, and torque. While machine learning (ML) approaches have been employed to predict joint kinematics from surface electromyography (sEMG) data, traditional ML models often struggle to generalize across dynamic movements. In contrast, physics-informed neural networks integrate biomechanical principles, but their effectiveness in predicting full movement dynamics has not been thoroughly explored. To address this, we introduce the Physics-informed Gated Recurrent Network (PiGRN), a novel model designed to predict multi-joint movement dynamics from sEMG data. PiGRN uses a Gated Recurrent Unit (GRU) to process time-series sEMG inputs, estimate multi-joint kinematics and external loads, and predict joint torque while incorporating physics-based constraints during training. Experimental validation, using sEMG data from five participants performing elbow flexion-extension tasks with 0 kg, 2 kg, and 4 kg loads, showed that PiGRN accurately predicted joint torques for 10 novel movements. RMSE values ranged from 4.02\% to 11.40\%, with correlation coefficients between 0.87 and 0.98. These results underscore PiGRN's potential for real-time applications in exoskeletons and rehabilitation. Future work will focus on expanding datasets, improving musculoskeletal models, and investigating unsupervised learning approaches.
\end{abstract}

\keywords{Surface electromyography \and Joint dynamics \and Human machine interface\and Gated Recurrent Unit, Physics-informed neural network \and Inverse dynamics}

\section{Introduction}
In recent decades, wearable robotic assistive devices have significantly evolved, boosting human power and aiding rehabilitation \cite{xia2024shaping}. Upper extremity exoskeletons, for instance, enhance strength and endurance, enabling users to exceed their usual capabilities, making them promising for assisting personnel in heavy item transport and aiding elderly and disabled individuals in rehabilitation \cite{kapsalyamov2020state, zhang2021development}. A vital aspect of these devices is the human-machine interface (HMI), which aims to understand human behavior for effective control \cite{esposito2021biosignal}, by accurately detecting human movement dynamics. This allows devices to respond naturally to the user's movements \cite{kwon2011real}. Modeling human movement dynamics involves parameters such as joint angles, angular velocities, angular accelerations, joint forces, and torques. While direct sensing and tracking of these parameters using sensors can be effective, it may introduce delays and limited flexibility for assistive devices \cite{young2016state, tucker2015control}, as these methods detect movements only after they occur, failing to provide real-time assistance.

Surface electromyography (sEMG) shows promise as a natural HMI, measuring neuromuscular electrical potentials noninvasively and predicting movement dynamics beforehand \cite{zheng2022surface}. Physics-based neuromusculoskeletal (NMS) models have used sEMG to develop effective exoskeleton control strategies based on joint torques \cite{durandau2019voluntary,pizzolato2015ceinms,lloyd2003emg}. However, challenges such as complex muscle geometry modeling and the need for intricate calibration limit its real-time applicability in assistive device control.

Researchers are using artificial intelligence to link sEMG signals with joint movement dynamics, employing regression-based machine learning. Neural networks, like dynamic RNNs and ANNs, have shown effectiveness in analyzing muscle activity and limb movement, predicting various parameters from sEMG signals, such as muscle force, limb kinematics, and kinetics \cite{cheron1996dynamic, liu1999dynamic, au2000emg, subasi2006classification, hahn2007feasibility, song2005using}. Research has further refined these methods for better sEMG-to-joint angle mapping and developing virtual human models \cite{chandrapal2011investigating,aung2013estimation,ngeo2014continuous}. Recent advancements also include the exploration of individual deep learning models such as CNNs, RNNs, autoencoders, and LSTMs, as well as their combinations, for estimating limb movement dynamics from multi-channel EMG signals \cite{xia2018emg,chen2019continuous,huang2020joint,zhang2022lower}. Training complex deep learning models can be a time-consuming process, particularly when dealing with large datasets, leading to substantial costs. Additionally, pre-trained models frequently do not perform well when applied to new subjects' data \cite{solares2020deep}. Transfer learning is suggested for tasks like predicting muscle forces and estimating limb kinematics for new subjects with limited data \cite{kim2019subject,zhang2022boosting}. However, these deep learning methods tend to act as 'black boxes,' often ignoring the underlying physics of neuromusculoskeletal processes.

Advancements in musculoskeletal (MSK) modeling and joint kinematics employ physics-informed machine learning techniques, enhancing flexibility and computational efficiency while adhering to physical principles \cite{karniadakis2021physics,meng2022physics}. One study integrated MSK modeling into a CNN-based physics-informed model to map EMG signals to muscle forces and joint kinematics \cite{zhang2022physics}. Zhang et al. further personalized the MSK model using CNN-based transfer learning and reduced training time through a distributed framework \cite{zhang2022boosting,zhang2023towards}. Taneja et al. used an ANN-based physics-informed parameter identification neural network to predict motion and identify parameters in MSK systems from raw sEMG signals and joint motion data \cite{taneja2022feature}. Another study combined MSK physics principles with a GAN to estimate muscle force and joint kinematics from sEMG signals \cite{shi2023physics}. Ma et al. proposed an ANN-based physics-informed neural network approach to predict muscle forces and identify muscle-tendon parameters using unlabeled sEMG data \cite{ma2024physics}. Despite these advancements, few challenges remain in integrating physics and machine learning in MSK modeling.

The above mentioned studies utilized deep learning models based on ANN or CNN, which have a limitation: they require fixed input sizes, posing challenges for real-time applications \cite{zhang2022physics, zhang2022boosting, zhang2023towards, ma2024physics}. sEMG signals, being time-series data with variable lengths, conflict with this requirement, making ANN or CNN models unsuitable for dynamic input data structures. Muscle force prediction was their key focus, often determined using surrogate models like the CMC tool from OpenSim \cite{delp2007opensim}. While useful for computing ground-truth values, surrogate models have drawbacks. They may not provide precise label data, leading to erroneous predictions, and their modeling and simulation can be costly and time-consuming, along with challenges in data recording. Additionally, most studies focused on single-joint movements, neglecting the transferability of models trained on sEMG data across different muscle loading scenarios. Understanding how variations in EMG signal strength affect model performance is crucial for diverse loading scenarios.

In recent years, the Gated Recurrent Unit (GRU), introduced by Kyunghyun Cho et al. in 2014 \cite{cho2014learning}, offers a simplified alternative to traditional RNNs, resembling LSTM networks \cite{chung2014empirical}. GRUs streamline structure reduces parameter count while maintaining performance. They are proficient in analyzing time series data, GRUs excel in capturing long dependencies without vanishing gradient issues \cite{noh2021analysis,nosouhian2021review}. Furthermore, their recursive design enables them to sustain and update a hidden state across time, efficiently handling sequential data and capturing temporal patterns \cite{shewalkar2019performance}. Hence GRUs are ideal for processing sEMG signal time series, representing muscle activity patterns effectively.
\clearpage
This work aims to achieve the following objectives:
\begin{enumerate}
    \item[I] \textbf{Predicting joint dynamics using a Physics-Informed Gated Recurrent Network (PiGRN) model:} We developed a PiGRN model by integrating a Gated Recurrent Unit (GRU) with physics-based constraints to process time-series sEMG data and predict multi-joint movement dynamics parameters, such as joint angle, velocity, acceleration, and external mass. Which is further used to estimate joint torque.

    \item[II] \textbf{Experimental validation of PiGRN for joint dynamics prediction:} To assess the performance of PiGRN, we have collected sEMG data from five participants performing elbow flexion-extension tasks under varying external loads (0 kg, 2 kg, and 4 kg).
    
    \item[III] \textbf{Generalization across multiple joints and load:} We validated the robustness of the PiGRN model in predicting multi-joint dynamics across varied muscle loading scenarios in an upper limb model.
\end{enumerate}

The rest of this article proceeds as follows: Section II details the methodology, including GRU architecture, inverse dynamics equations, PiGRN architecture, and combined loss function design. Section III explains Systems Validation, covering experimental setup, data processing, feeding, and training, along with PiGRN model hyperparameters, torque prediction, and evaluation criteria. Section IV presents results, while Section V discusses future work, and Section VI concludes.

\section{Methodology}
\subsection {The Gated Recurrent Unit (GRU) Architecture}
\begin{figure}[ht!]
\centerline{\includegraphics[trim=0.5cm 1.5cm 0.5cm 2.0cm, clip=true,width=0.6\linewidth]
{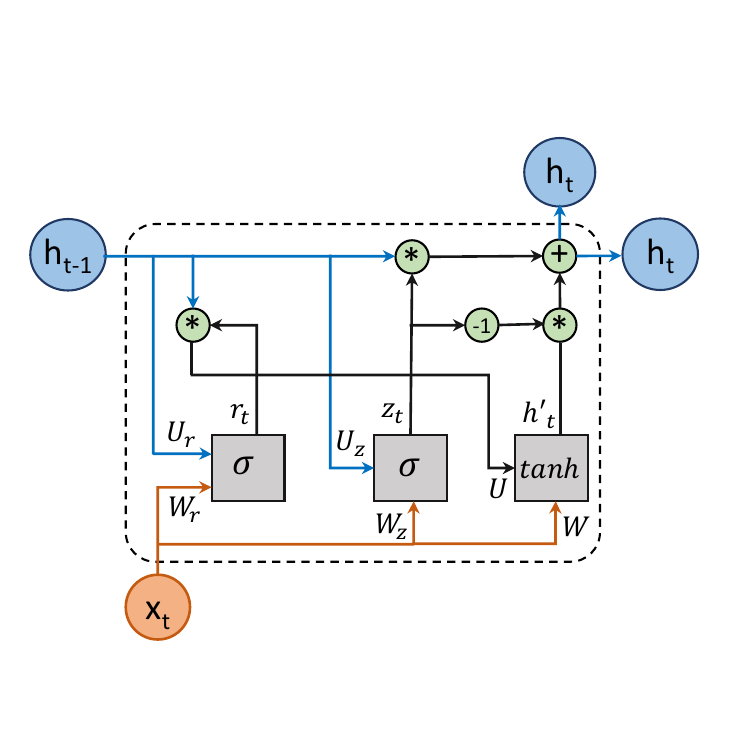}}
\caption{The GRU framework}
\label{Fig:GRU}
\end{figure}

\begin{figure}[ht!]
\centerline{\includegraphics[trim=0.5cm 3.0cm 0.5cm 1.65cm, clip=true,width=0.6\linewidth]
{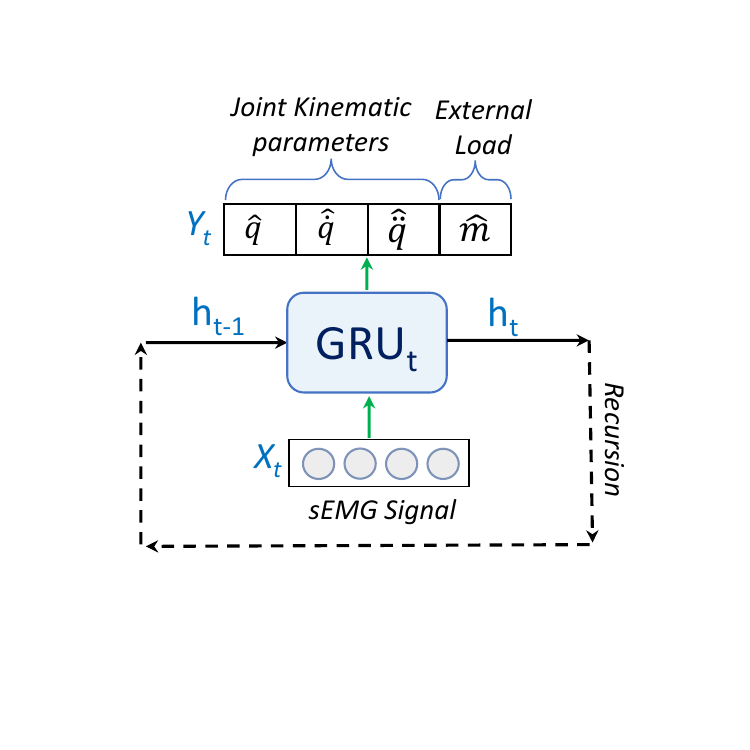}}
\caption{The recursive model of GRU at time instance \(t\)}
\label{Fig:Recursion}
\end{figure}

The Figure \ref{Fig:GRU} illustrates the architecture of a Gated Recurrent Unit (GRU), comprising two gates: an update gate ($z_t$) and a reset gate ($r_t$). These gates play a crucial role in regulating the flow of information within the network. Within this framework, as depicted in Figure. \ref{Fig:Recursion}, the recursive nature enables the model to effectively retain and update information across sequential sEMG data. At each time step, the GRU receives an input sEMG vector $x_t$ and combines it with the previous hidden state $h_{t-1}$ to compute an intermediate state. The update gate determines how much of the previous memory to retain by computing $z_t$ using $h_{t-1}$, the previous hidden state, and $x_t$ the current input. The reset gate decides how much of the past information to forget using $r_t$. The candidate hidden state $h'_{t}$ is then computed using $r_t$, $h_{t-1}$, and $x_t$. Finally, the new hidden state $h_t$ combines the previous hidden state $h_{t-1}$ and the candidate hidden state $h'_{t}$ to predict joint kinematic parameters and external load sequence. Here, the key equations that govern the information flow within a GRU is explained.
\vspace{-3mm}
\[ Update \hspace{0.1cm} Gate: z_t = \sigma\*(W_z*x_{t} + U_z*h_{t-1}) \]
\[Reset \hspace{0.1cm} Gate: r_t = \sigma\*(W_r*x_{t} + U_r*h_{t-1}) \]
\[ Candidate \hspace{0.1cm}Hidden\hspace{0.1cm}State: h'_{t} = \tanh\*(W*x_{t} + U*r_t*h_{t-1}) \]
\[ Hidden \hspace{0.1cm} state: h_{t} = z_t*h_{t-1} + (1-z_t)*h'_{t} \]
\[ Output: \hat{y_{t}} = W^y*h_{t} \]
Here, \(W_i\) and \(U_i\) are the weight matrices corresponding to the current input \(x_t\) and the previous hidden state \(h_{t-1}\) respectively, with \(i\) being \(z\) and \(r\). The GRU architecture utilizes the sigmoid function, \(\sigma\), and the hyperbolic tangent function, \(\tanh\), as activation functions. Additionally, \(W\) is another weight matrix used in the computation of the candidate hidden state.
\subsection {Inverse Dynamics Equation of Motion}
This work utilizes the inverse dynamics equation of motion method to estimate joint torques in the musculoskeletal (MSK) systems. Specifically, we employ a previously derived equation of motion for elbow flexion-extension (FE) motion in the sagittal plane, considering an external load ($m$) held in the hand. The FE motion involves two degrees of freedom (DOF) at the shoulder and elbow joints, modeled as revolute joints. The MSK comprises the upper arm (humerus) and the lower arm (ulna and radius), treated as a rigid segments. Body segment mass and center of mass are estimated following methods described in Herman et al. \cite{herman2016physics}, while segment moment of inertia (MOI) is calculated using regression equations from Hinrichs et al. \cite{hinrichs1985regression}. The equation of motion is represented by Equation \ref{Equation of motion}.
\begin{equation}\label{Equation of motion}
    I(q(t),m) {\ddot{q}(t)}+{C((q(t),m)}, {\dot{q}(t)})+{{G(q(t),m})} = {\tau(t)}
\end{equation}
where $q(t)$, ${\dot{q}}(t)$, ${\ddot{q}}(t)$ are the joint kinematics parameters including angle, angular velocities, and angular accelerations, respectively; $I(q(t),m)$ is the system inertia matrix; $C((q(t),m),{\dot{q(t)}})$ is the centrifugal and Coriolis force matrix; $G(q(t),m)$ gravitational force on the system all as a function of external load ($m$).
\subsection {Main Framework of Physics Informed Gated Recurrent Network (PiGRN)}
\begin{figure*}[ht!]
\centerline{\includegraphics[trim=0.12cm 2.3cm 0.2cm 1.15cm, clip=true,width=\linewidth]
{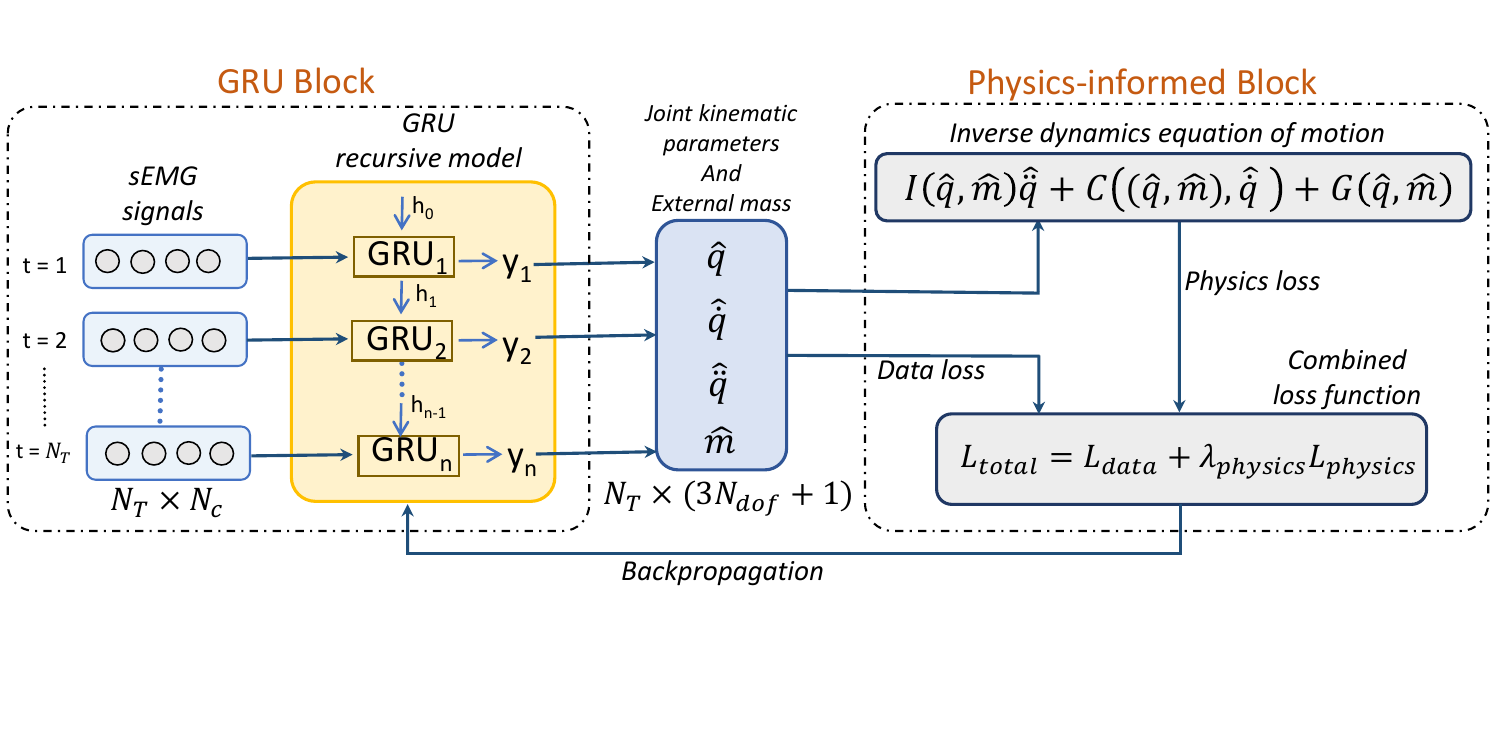}}
\caption{Physics Informed Gated Recurrent Network Architecture (PiGRN), here, $N_T$ denotes length of time series data, $t$ denotes time instance. $N_C$ and $N_{dof}$ denotes number of channels of sEMG signal and total number of degrees of freedom in the MSK system , $h_{i-1}$ and $GRU_t$ denotes hidden state and GRU block at at time instance $t$. $\lambda_physics$ is physics loss weighting factor which is a hyperparameter. Losses are based on MSE i.e. Mean square error.}
\label{Fig:PiGRN}
\end{figure*}

Figure \ref{Fig:PiGRN} illustrates the main framework of the proposed method for training the Physics Informed Gated Recurrent Network (PiGRN) using varied load sEMG signals. The GRU takes sEMG signals as input and predicts joint kinematic parameters, including joint angles ($q(t)$), angular velocities (${\dot{q}}(t)$), and angular accelerations (${\ddot{q}}(t)$) for all joints and external load ($m$) in the musculoskeletal (MSK) model.
Here a GRU model is employed to extract temporal dependencies in muscle activation patterns and predict joint kinematics from time-series EMG data. Initially, sEMG signals recorded under varied loads are input into the GRU, which then predicts joint kinematics. These predictions are then used in the MSK equation to calculate predicted torque, compared with actual torque to determine physics loss. This physics loss is incorporated into the GRU's loss function, ensuring a combined loss that minimizes both data loss and physics-based loss. Ultimately, this personalized network effectively maps sEMG signals to joint kinematics.
\subsection {Design of Combined Loss Function}
To train the proposed PiGRN, the framework utilizes a combined loss function comprising two main components: the standard data loss (representing supervised learning) and the physics-based loss (representing residual loss). The data loss, calculated as Mean Squared Error (MSE) loss, aims to minimize the discrepancy between the ground truth and the predicted values. On the other hand, the physics-based loss ensures adherence to physical constraints governed by the equation of motion, achieving this by minimizing the residual of joint torque. The components of the loss functions in PiGRN are as follows:
The data loss (supervised learning loss):
\begin{equation}\label{Equation data loss}
L_{\text{data}} = L_{q} + L_{\dot{q}} + L_{\ddot{q}} + L_{m}
\end{equation}
where, the total data loss, denoted as \( L_{\text{data}} \), is obtained by summing \(L_{q}\), \(L_{\dot{q}}\), \(L_{\ddot{q}}\), and \(L_{m}\), which represents the losses for joint angle, angular velocity, angular acceleration, and external load respectively. Each of these data loss functions is defined as the mean squared error between the predicted and target values, as given in Equation \ref{Equation data loss dummy}.
\begin{equation}\label{Equation data loss dummy}
L_{\text{u}} = \frac{1}{N} \sum_{i=1}^{N} \left|(\hat{ u }(x_i); {\hat{\theta}}) - u_i \right|^2
\end{equation}
where \( \mathcal{D} = \{(x_i, u_i)\}_{i=1}^{N} \) is a set of measured data points with \( x_i \) as sEMG inputs and \( u_i \) as corresponding joint kinematics values, and \( \hat{u}(x_i) \) is the network's prediction at \( x_i \) with trainable parameters $\hat{\theta}$.
The physics loss (Residual Loss):
\begin{equation}\label{Equation physics loss}
L_{\text{physics}} = \frac{1}{N} \sum_{i=1}^{N} \left| \mathcal{F}((\hat{ q }(x_i), \hat{ \dot{q} }(x_i), \hat{ \ddot{q} }(x_i), \hat{ m }(x_i)); {\hat{\theta}})\right|^2
\end{equation}
where \( \mathcal{F} \) represents the residual of the joint torque, \( x_i \) are sEMG data points, and \( N \) is the number of data points.
The combined loss function:
The total loss function \( L_{\text{total}} \) in a PiGRN is a weighted sum of these individual loss components:
\begin{equation}\label{Equation combined loss}
L_{\text{total}} = \lambda_{\text{data}} L_{\text{data}} + \lambda_{\text{physics}} L_{\text{physics}}
\end{equation}
where \( \lambda_{\text{data}} \), and \( \lambda_{\text{physics}} \) are hyperparameters controlling the relative importance of each term, here the value of \( \lambda_{\text{data}} \) is one.

The neural network is trained by minimizing the total loss function \( L_{\text{total}} \), Equation \ref{Equation combined loss}. This involves adjusting the network parameters $\hat{\theta}$ so that the predicted solution \((\hat{ q }(x_i); {\hat{\theta}}), (\hat{ \dot{q} }(x_i); {\hat{\theta}}), (\hat{ \ddot{q} }(x_i) ); {\hat{\theta}})\), and \((\hat{ m }(x_i); {\hat{\theta}})\) fits the observed data, while also adhering to the physical law represented by the equation of motion of the MSK system dynamics during movements as described in Equation \ref{equation_tau_residual}.
\begin{equation}\label{equation_tau_residual}
\begin{split}
\mathcal{F}(\hat{q}_i, \hat{\dot{q}}_i, \hat{\ddot{q}}_i, \hat{m}_i; {\hat{\theta}}) = 
I(\hat{q}_i,\hat{m}_i;\hat{\theta}) \hat{\ddot{q}}_i + C((\hat{q}_i,\hat{m}_i;\hat{\theta}), \hat{\dot{q}}_i) \hdots \\ \hdots
+ G(\hat{q}_i,\hat{m}_i;\hat{\theta}) - \tau
\end{split}
\end{equation}
Here, $\mathcal{F}(\hat{ q }_i, \hat{ \dot{q} }_i, \hat{ \ddot{q} }_i, \hat{ m }_i ; {\hat{\theta}})$, represents residuals related to Equation \ref{equation_tau_residual} for the $i^{th}$ sample. \((\hat{q}(x_i); \hat{\theta}\)), \((\hat{\dot{q}}(x_i); \hat{\theta}\)), \((\hat{\ddot{q}}(x_i); \hat{\theta}\)), and \((\hat{m}(x_i); \hat{\theta}\)) refers to the joint kinematics and the external load predicted by the GRU model with trainable parameters $\hat{\theta}$ for the $i^{th}$ sample. The most accurate predictions are achieved by minimizing the combined loss function $J$, as defined in Equation \ref{equation min loss}, for the optimal GRU parameters denoted as ${\theta}$.
\begin{equation}\label{equation min loss}
{\theta} = \underset{\hat{\theta}}{\operatorname{argmin}}\,(J), \\
\quad \text{where} \; J = L_{\text{data}} + \lambda_{\text{physics}} L_{\text{physics}}
\end{equation}
\section{SYSTEM VALIDATION PROCEDURES}
\subsection {Experimental setup, Data collection and Data processing}
\subsubsection{Experimental setup}
The study enlisted five healthy male volunteers, aged $26 \pm 3$ years, with an average body mass of $73 \pm 4$ kg and a height of $174 \pm 3.5$ cm. 
\begin{figure*}[h]
\centerline{\includegraphics[trim=0.1cm 3.0cm 0.18cm 1.1cm, clip=true,width=0.98\linewidth]
{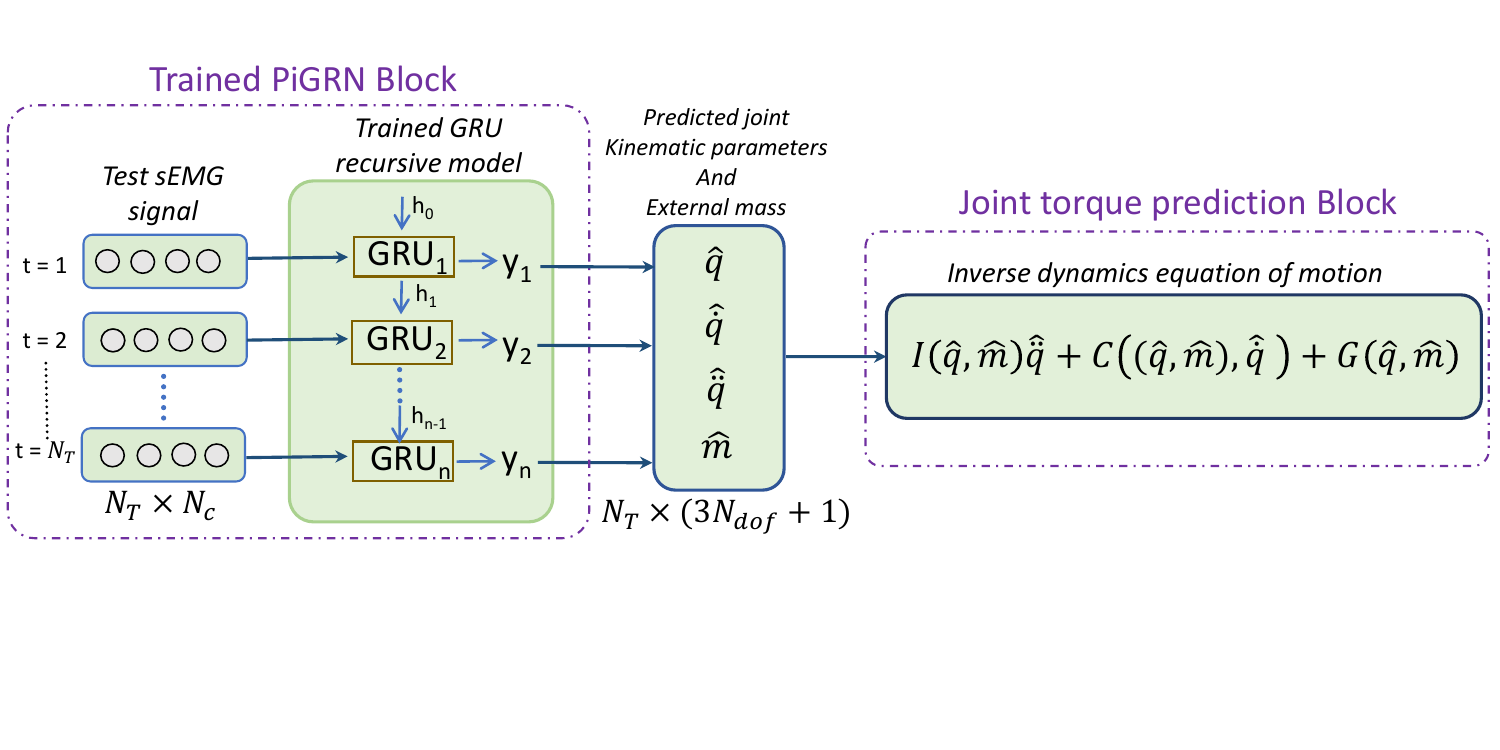}}
\caption{Joint torque estimation flow chart once the PiGRN model is trained}
\label{Fig:Trained PiGRN}
\end{figure*}
These participants engaged in elbow flexion-extension (FE) exercises utilizing 0 kg, 2 kg, and 4 kg weights, enabling movement along the sagittal plane for both the elbow and shoulder joints. Concurrent measurements of surface electromyography (sEMG) signals and joint angle data were recorded. Approval for the experimental procedure was obtained from the All India Institute of Medical Sciences (AIIMS) in New Delhi. The sEMG data were captured using Noraxon’s Ultium EMG sensor system (4000 Hz), while the elbow joint angle was tracked employing the Noraxon Ninox camera system (125 Hz).
\subsubsection{Data collecion}
The study recorded sEMG signals and joint angles from five participants. Four electrodes were placed on each participant's biceps long head, biceps short head, triceps long head, and triceps lateral head muscles, following guidelines from \cite{konrad2005abc}. Additionally, four reflective markers tracked shoulder and elbow joint angles. Each trial began with a screen cross and beep, signaling motion initiation, and ended manually. A two-second rest followed each elbow flexion-extension (FE). Participants completed 12 runs, with 4 runs each for 0 kg, 2 kg, and 4 kg loads, comprising 10 FE trials per run.
\subsubsection{Data pre-processing} 
We enhanced joint angle data by applying a Gaussian filter with a 10-standard deviation and a 6-window size. For sEMG data, we first filtered signals between 10-450 Hz to remove noise, then converted negative amplitudes to positive through full-wave rectification. Next, a 7 Hz low-pass filter smoothed the signal, yielding linear envelopes. These were normalized by dividing them by the maximum voluntary contraction (MVC) EMG values from recorded trials, ensuring consistency across trials and subjects. Given EMG signals' 4000 Hz frequency and joint angle recordings' 125 Hz, we downsampled the EMG envelope by 32 to match joint angle data points.
\subsection{Data feeding and Training of PiGRN}
The dataset was divided into a 3:1 ratio for training and testing. Angular velocity and acceleration were computed from joint angles using the central difference method. Data was collected in sets of 10 elbow flexion-extension (FE) movements, with rest between each repetition. To speed up training and simplify data handling, we segmented the data into complete FE movement repetitions, each comprising 800 timesteps, including rest. Without rest, each FE repetition was approximately 450 timesteps long. The GRU network takes sEMG data of size (800, 4, N), where 4 is the number of EMG channels, 800 is the timesteps, and N is the number of training datasets. These datasets were stacked and fed into the GRU model for training. The GRU model's output includes joint kinematics and external load, with dimensions (800, 7, N), representing elbow and shoulder joint angles, velocities, accelerations, and external load. These predicted values are then used as inputs for the inverse dynamics equations of motion to predict joint torques, resulting in dimensions (800, 2, N) for elbow and shoulder torques.
In a single training loop, the GRU iterates through each time step of an sEMG input sequence. It computes update and reset gates, calculates the candidate hidden state, and combines them to determine the current hidden state. These hidden states are then used to generate output, such as joint kinematic parameters and external load. Following this, the loss is computed by comparing the model's output with the actual targets. Through backpropagation through time, gradients of the loss with respect to GRU parameters are calculated, considering time step dependencies. These gradients update the GRU parameters using the Adam optimization algorithm. This loop repeats over multiple epochs and batches to train the GRU model effectively.
\subsection {Hyper Parameters of Model }
In our model, hyperparameter tuning was crucial. We used a linear/grid search approach to optimize parameters. The model comprises two stacked GRU cells processing EMG signals. We found that using 64 neurons in each hidden layer yielded optimal results. Adam optimization with a learning rate of 0.0001 was chosen for faster convergence. To prevent overfitting, we applied a dropout probability of 0.2 and introduced a physics loss weighting factor of 0.001, determined via linear search. We set the number of epochs to 2000, noting convergence after 1500 epochs due to Adam optimizer's efficiency.
\subsection {Joint torque prediction and evaluation criteria}
In this proposed framework, our main focus is to predict joint torque. After completing the training of the PiGRN model, the joint torque computation block is connected with the trained PiGRN block (see Fig. \ref{Fig:Trained PiGRN}). To predict joint torque, testing sEMG data is passed through the trained model to predict joint kinematic parameters and external mass. These predicted parameters are then used as inputs to the inverse dynamic equation of motion to predict joint torques. To evaluate the performance of the model, Percentage root mean square error (RMSE) and the Pearson coefficient of correlation are used, as described in Equation \ref{RMSE} and Equation \ref{Correlation Coefficient}, respectively. \% rmse has been used to assess the relative quality of prediction based on its RMSE score.
\begin{equation}\label{RMSE}
    \text{RMSE} = \sqrt{\frac{1}{n} \sum_{i=1}^{n} (y_i - \hat{y}_i)^2}
\end{equation}

\begin{equation}
    \% RMSE = \frac{RMSE*100}{max(y_i)}
\end{equation}
Where \( n \) is the number of observations or data points, \( y_i \) represents the actual value (ground truth) for the \( i \)-th observation, \( \hat{y}_i \) is the predicted value for the \( i \)-th observation, and \( \sum \) is the summation symbol, indicating a sum over all \( n \) observations.
\begin{equation}\label{Correlation Coefficient}
    r = \frac{\sum_{i=1}^{n} (x_i - \bar{x})(y_i - \bar{y})}{\sqrt{\sum_{i=1}^{n} (x_i - \bar{x})^2 \sum_{i=1}^{n} (y_i - \bar{y})^2}}
\end{equation}
Where $n$ is the number of paired observations, $x_i$ represents the value of the $i$-th observation for variable $x$, $\bar{x}$ is the mean (average) value of $x$, $y_i$ is the value of the $i$-th observation for variable $y$, $\bar{y}$ is the mean (average) value of $y$, and $\sum$ is the summation symbol, indicating sum over all $n$ observations.

\section{RESULTS}
In this section, we evaluate the performance of the proposed framework for elbow and shoulder joints by comparing it with selected baseline methods. We begin by describing the training process of the proposed framework.
\begin{figure}[ht]
\centerline{\includegraphics[width = 0.6\linewidth]
{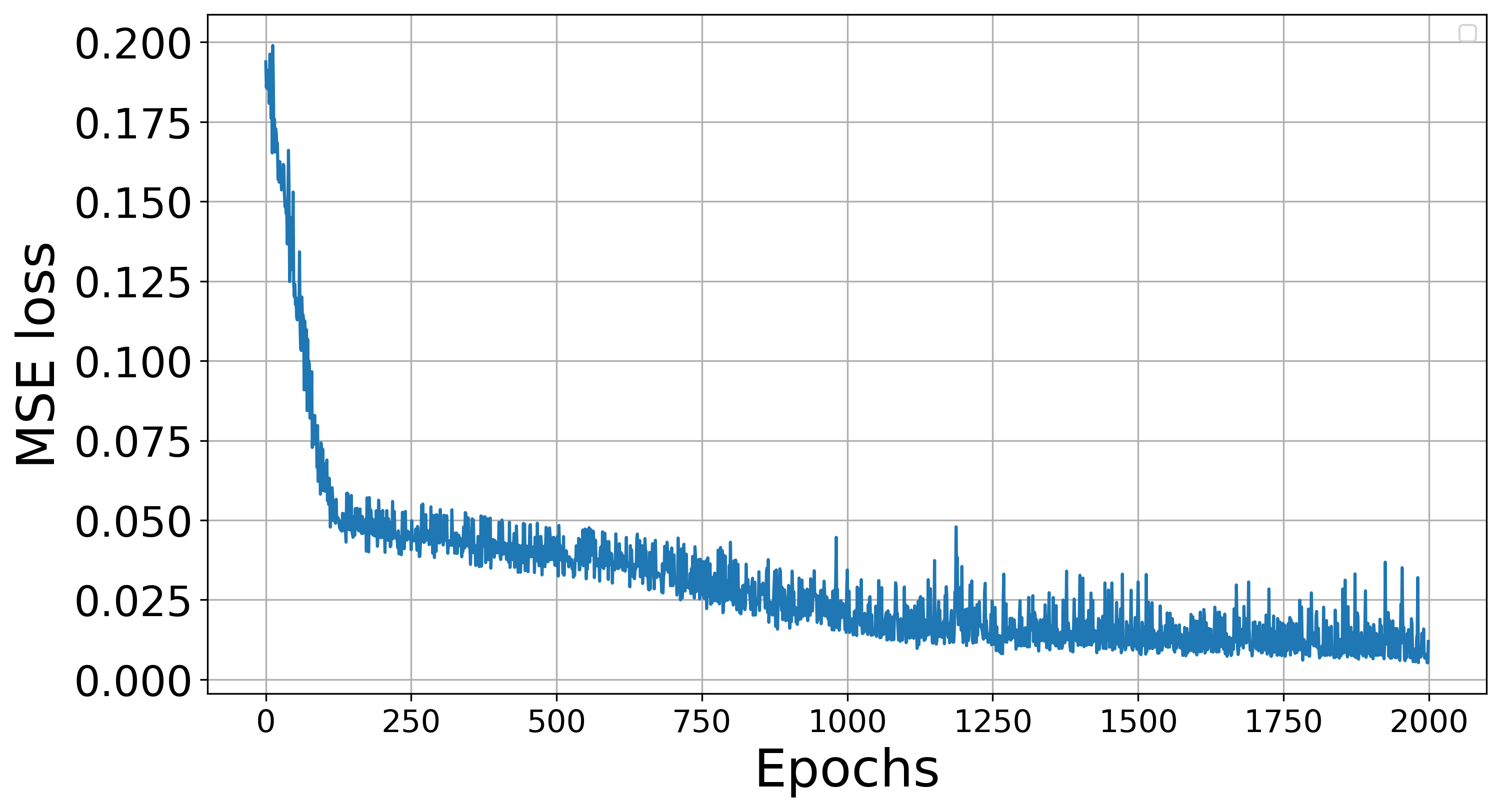}}
\caption{Loss curve during training}
\label{fig:loss curve}
\end{figure}
Subsequently, we conduct comparisons to showcase the predicted results of the proposed framework and the baseline methods, including representative outcomes for predicted joint torques, joint kinematic parameters, and external load, as well as detailed and average predictions for five healthy subjects. Finally, we consider the long sequence scenario to assess the robustness and generalization performance of the proposed framework.
\begin{figure*}[ht]
	\centering
	\begin{subfigure}{.31\linewidth}
		\includegraphics[width= \linewidth]{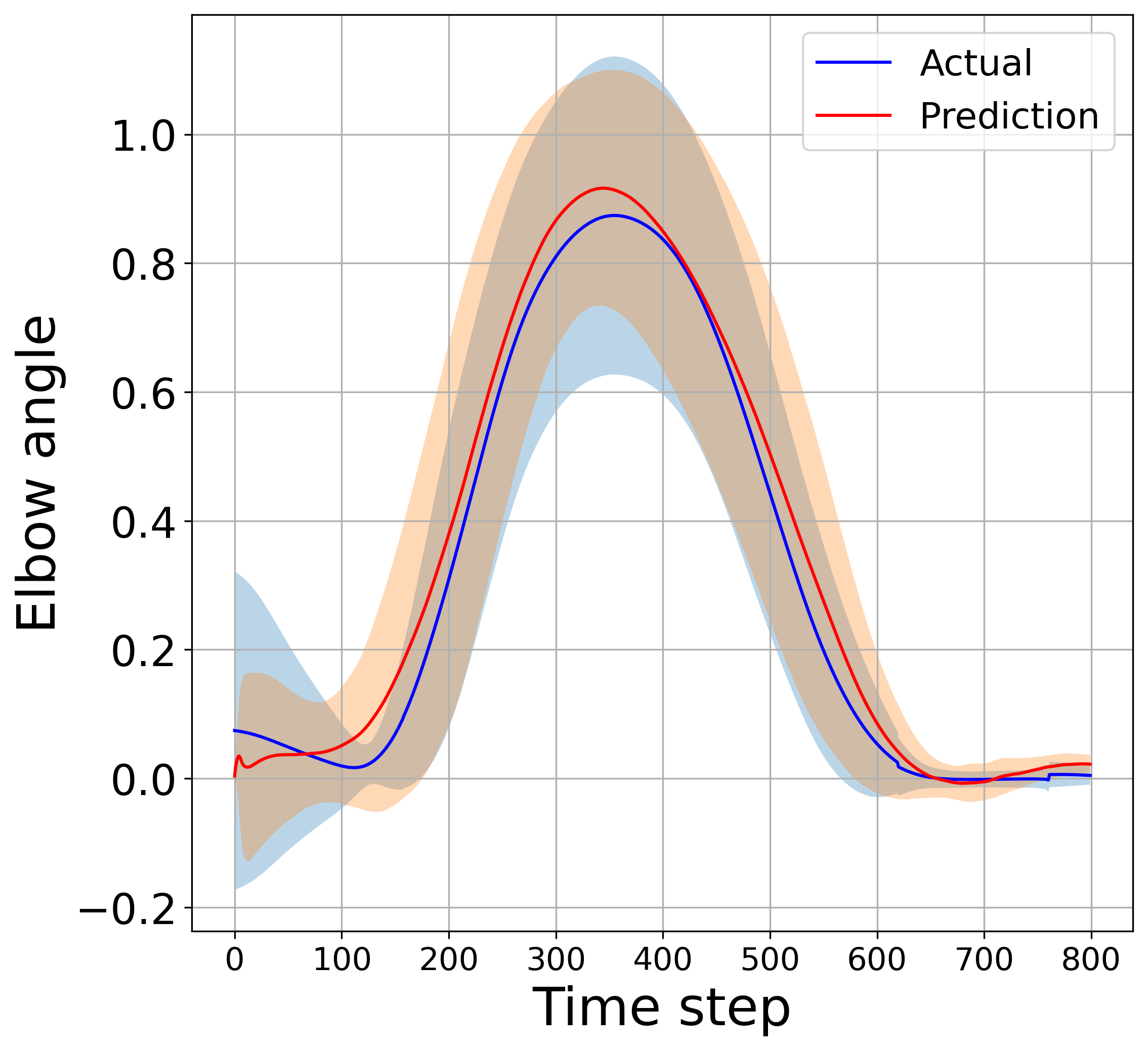}
		\caption{0KG load }
	\end{subfigure}
	\begin{subfigure}{0.30\linewidth}
		\includegraphics[width=\linewidth]{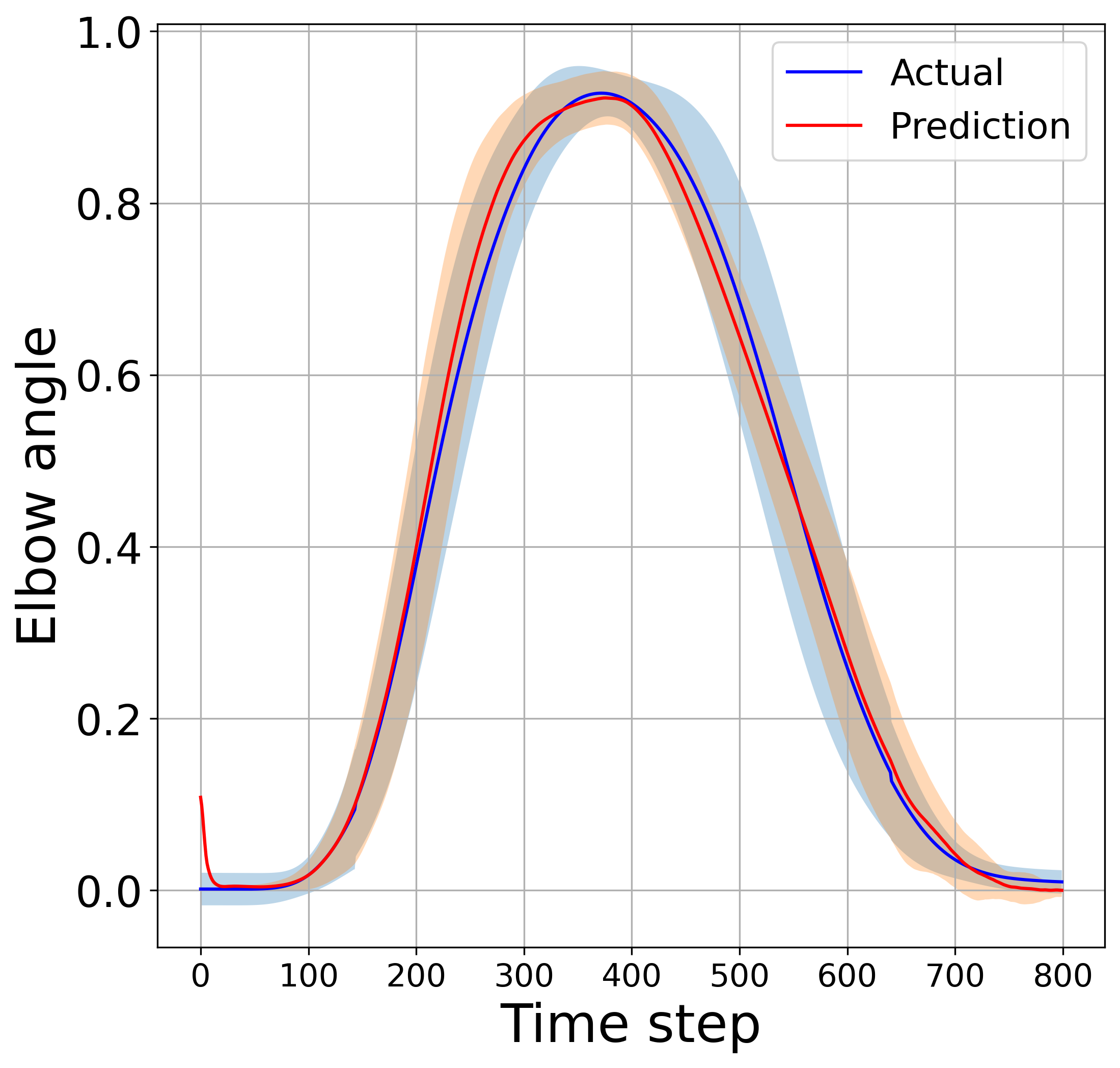}
		\caption{2KG load }
	\end{subfigure}
	\begin{subfigure}{0.3\linewidth}
		\includegraphics[width=\linewidth]{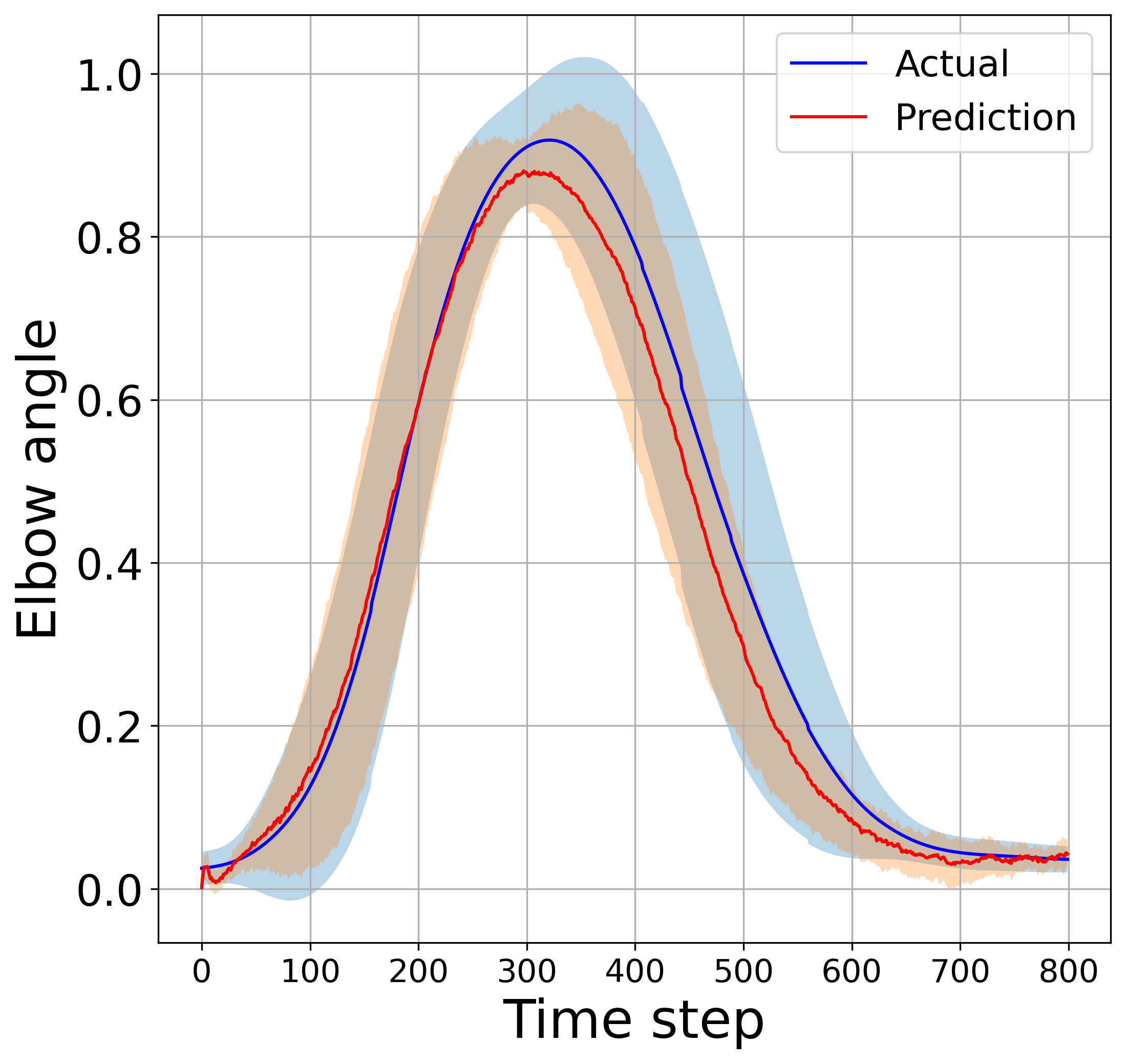}
		\caption{4KG load}
	\end{subfigure}
	\begin{subfigure}{0.315\linewidth}
		\includegraphics[width= \linewidth]{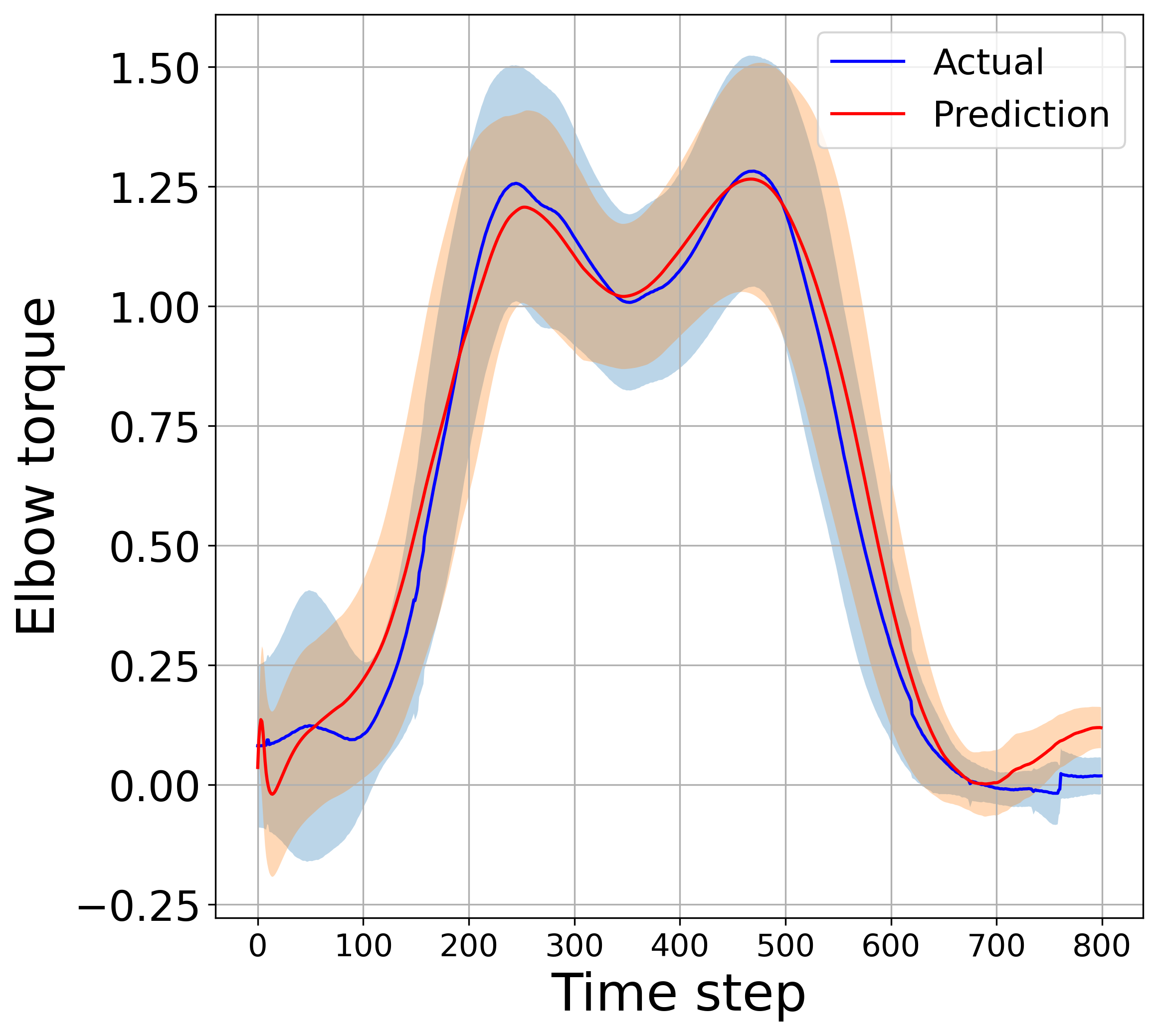}
		\caption{0KG load}
	\end{subfigure}
	\begin{subfigure}{0.29\textwidth}
		\includegraphics[width=\linewidth]{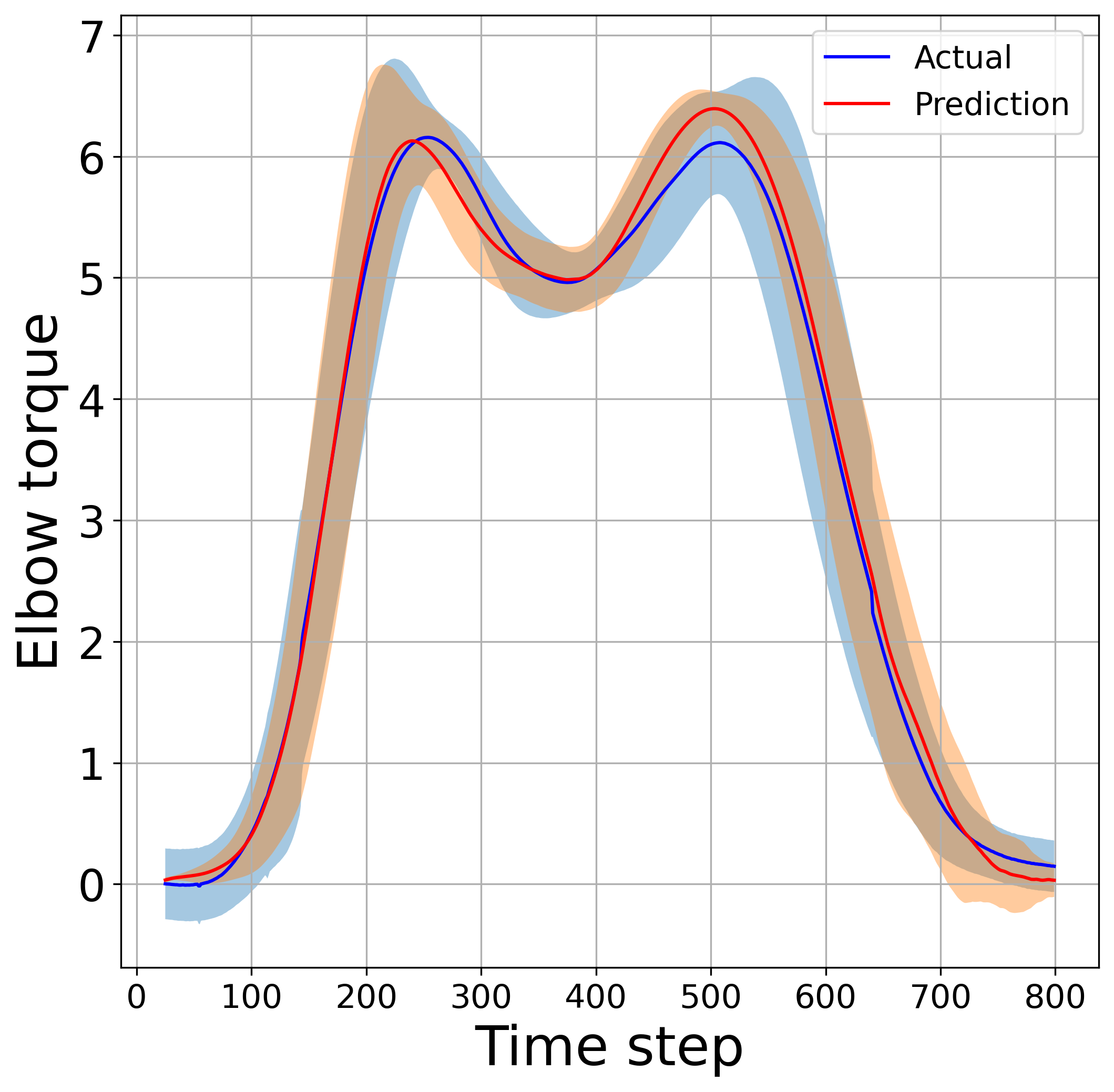}
		\caption{2KG load}
	\end{subfigure}
	\begin{subfigure}{0.3\linewidth}
		\includegraphics[width=\linewidth]{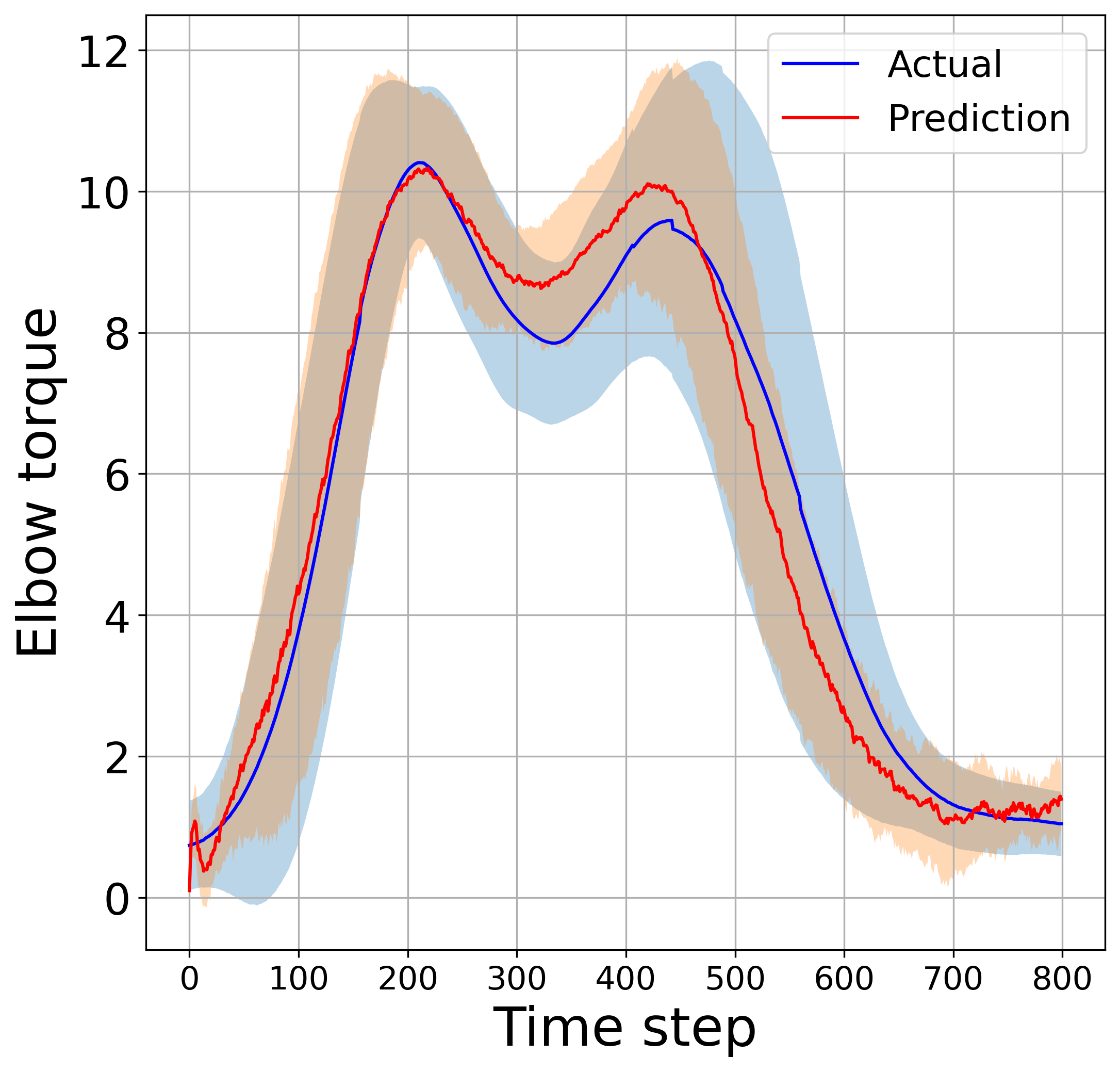}
		\caption{4KG load}
	\end{subfigure}
	\caption{Plots for Elbow joint angle and Elbow joint torque prediction of a subject at different loading at hand. Units of elbow torque is N-m where as elbow angles are in normalized in range (0-1) whose maximum value is about ${130}^0$.Solid line represents mean value of predicted $\&$ true value of all testing dataset whereas shaded region represents spread of them.}
    \label{fig: elbow angle and torque prediction}
\end{figure*}
\begin{figure*}[ht]
\begin{tikzpicture}
\begin{axis}[
    ybar,
    ylabel= {RMSE in percentage},
    height=9cm, width=17cm,
    ymax = 16,
    enlargelimits=0.1,
    legend style={at={(0.5,-0.12)},
	anchor=north,legend columns=-1},
    bar width=0.4cm,
    symbolic x coords={Subject 1, Subject 2, Subject 3, Subject 4, Subject 5, Mean},
    xtick = data,
    x tick label style={
		/pgf/number format/1000 sep=},
    nodes near coords,
    nodes near coords align={vertical},
    every node near coord/.append style={rotate=90, anchor=west},
    grid = both,
    ]
\addplot[fill = blue!60] coordinates {(Subject 1, 4.56) (Subject 2,4.21) (Subject 3, 3.5)(Subject 4, 4.01)(Subject 5, 4.6)(Mean,4.02)};
\addplot[fill = magenta!60] coordinates {(Subject 1, 7.98) (Subject 2,4.56) (Subject 3,5.6)(Subject 4, 10.0)(Subject 5, 7.57)(Mean,7.22)};
\addplot[fill = violet!70] coordinates {(Subject 1, 9.53) (Subject 2, 5.36) (Subject 3,15.6)(Subject 4, 12.44)(Subject 5, 14.43)(Mean,11.4)};
\addplot[fill = orange!70] coordinates {(Subject 1, 8.26) (Subject 2, 7.26) (Subject 3,14.56)(Subject 4, 12.44)(Subject 5, 15.6)(Mean,11.4)};
\legend{Elbow Angle , Elbow Torque , Shoulder Angle , Shoulder Torque}
\end{axis}
\end{tikzpicture}
\caption{RMSE scores for prediction of joint angle $\&$ torque for five different subjects using PiGRN model.}
\label{fig:RMSE scores of five subjects}
\end{figure*}
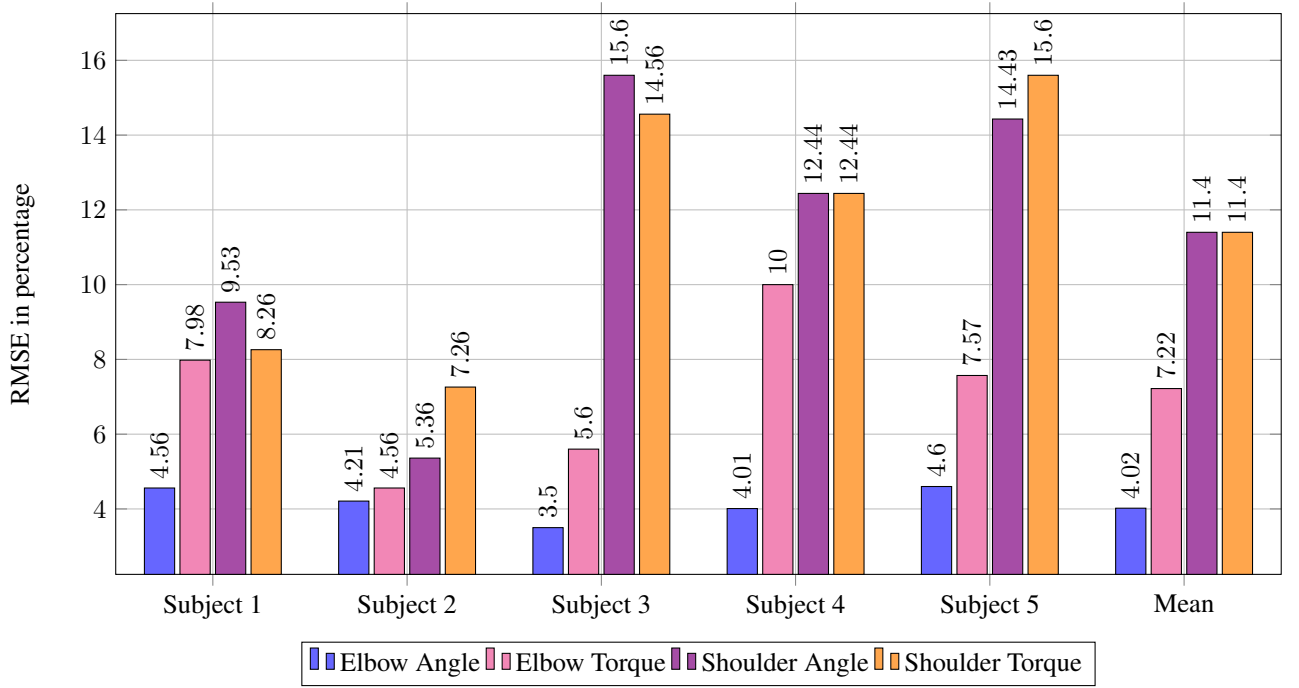
\begin{figure*}[ht]
	\centering
	\begin{subfigure}{.16\linewidth}
		\includegraphics[width= \linewidth]{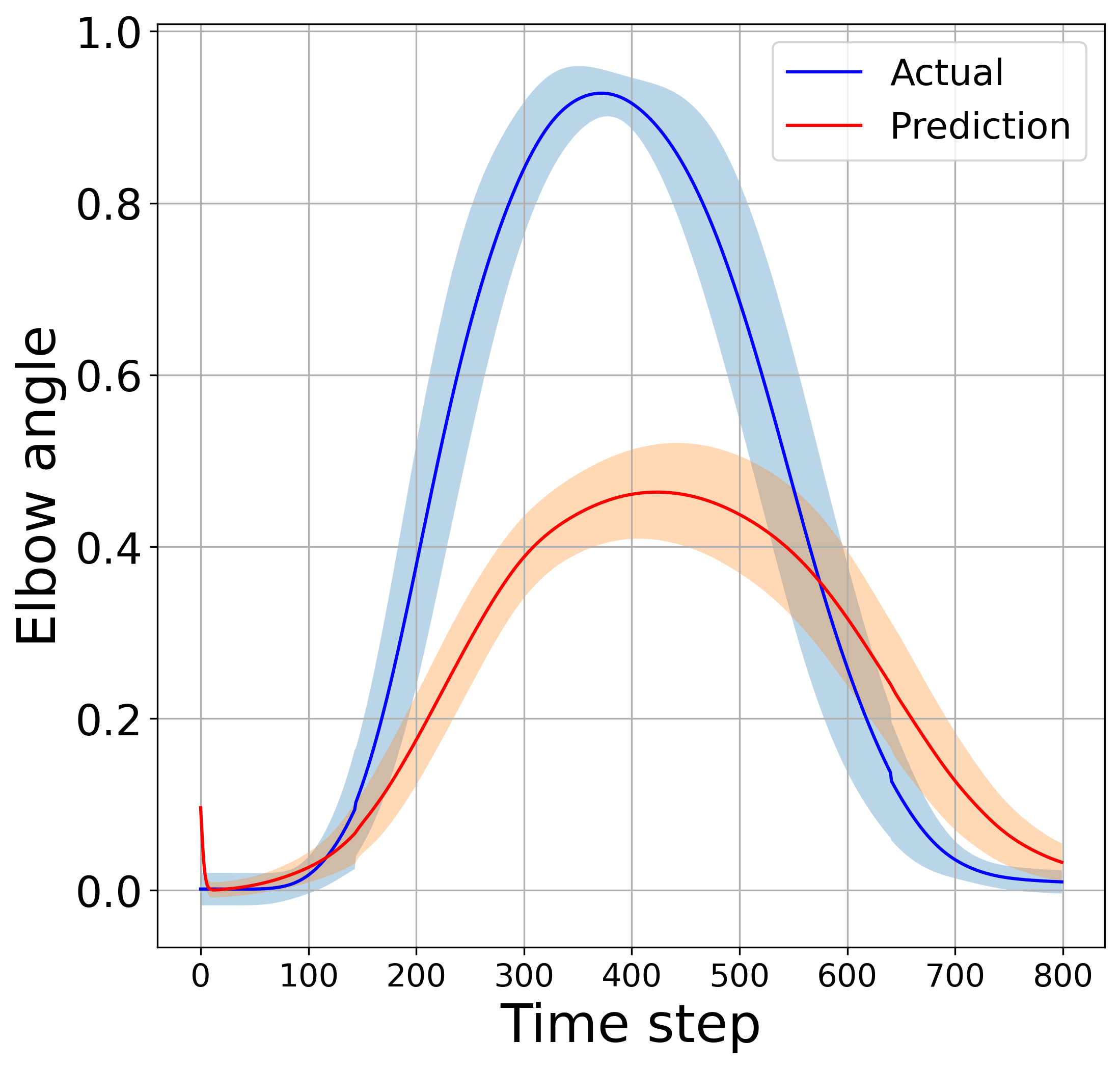}
		\caption{(1.0, 0.21)}
	\end{subfigure}
	\begin{subfigure}{.16\linewidth}
		\includegraphics[width=\linewidth]{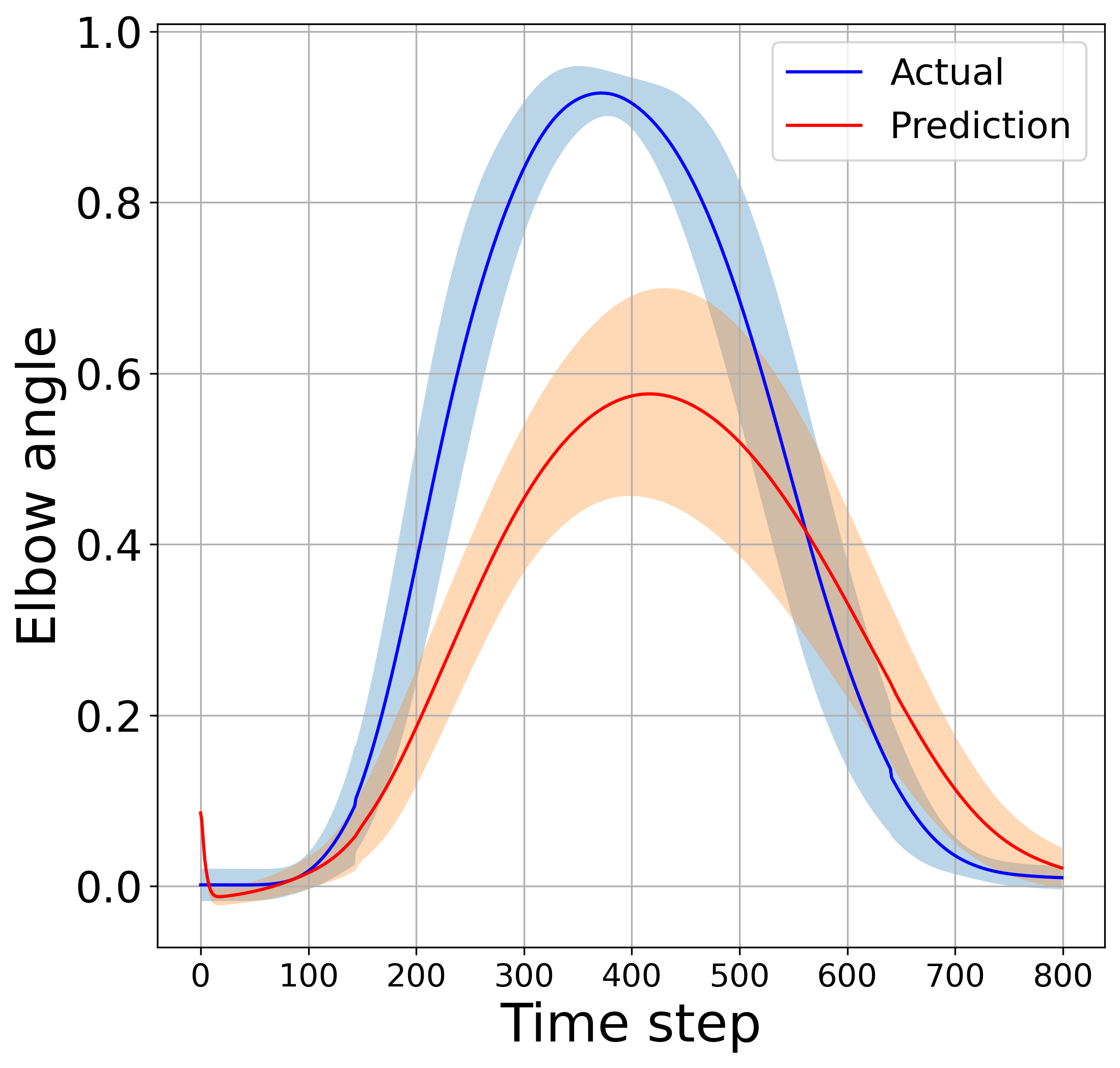}
		\caption{(0.1, 0.17)}
	\end{subfigure}
	\begin{subfigure}{.16\linewidth}
		\includegraphics[width=\linewidth]{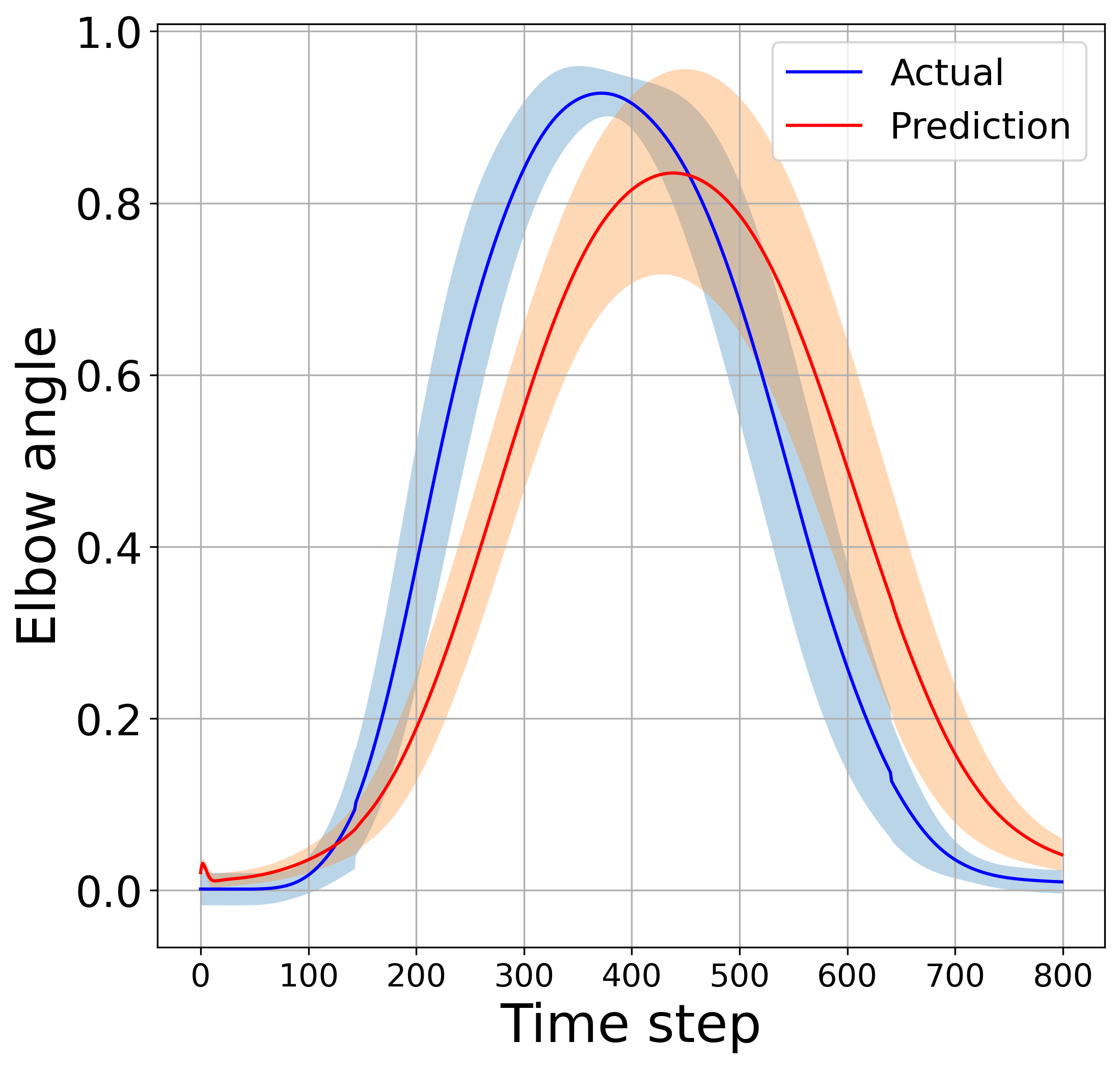}
		\caption{(0.01, 0.13)}
	\end{subfigure}
	\begin{subfigure}{.16\linewidth}
		\includegraphics[width= \linewidth]{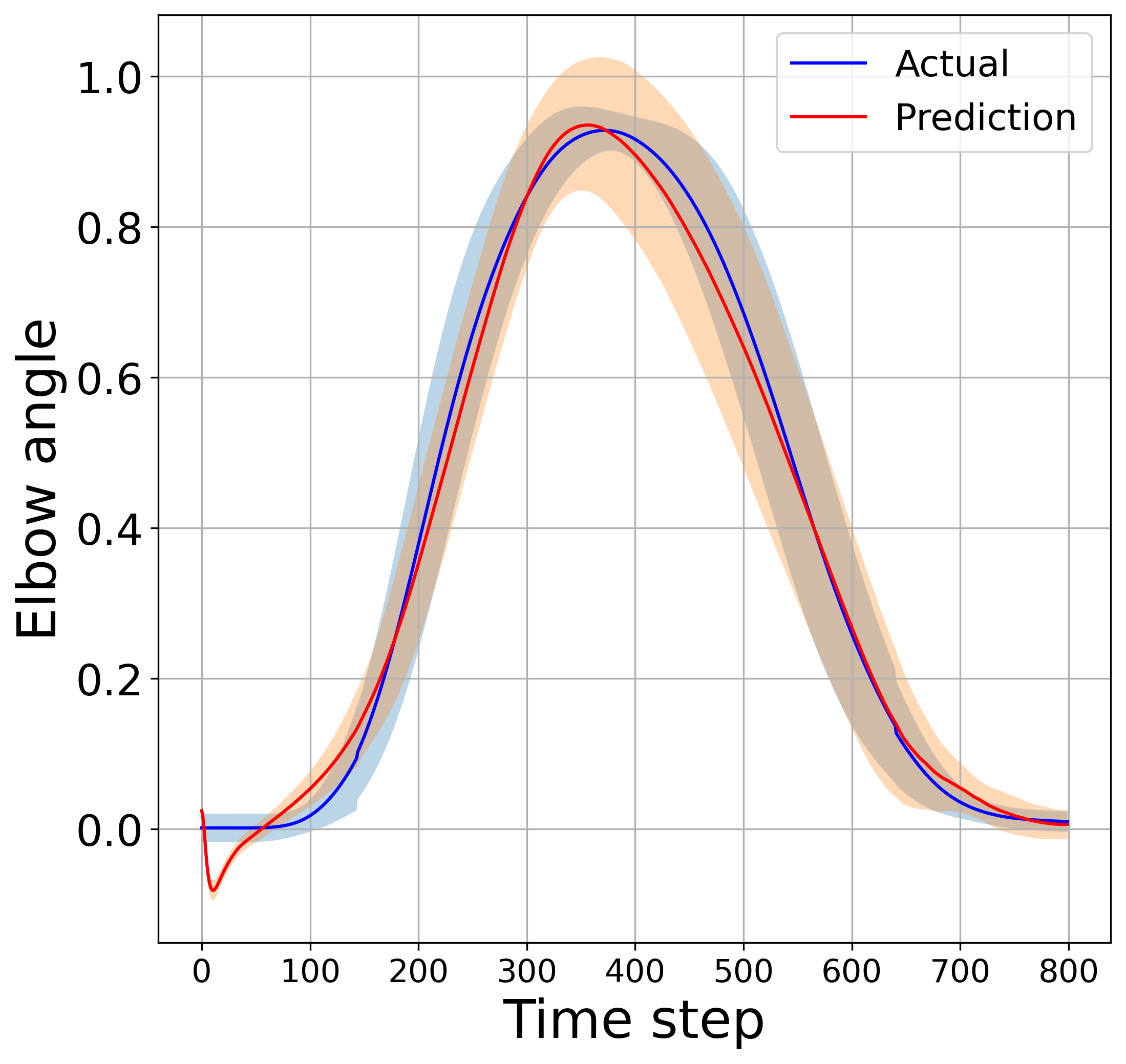}
		\caption{(0.05, 0.065)}
	\end{subfigure}
	\begin{subfigure}{.16\linewidth}
		\includegraphics[width=\linewidth]{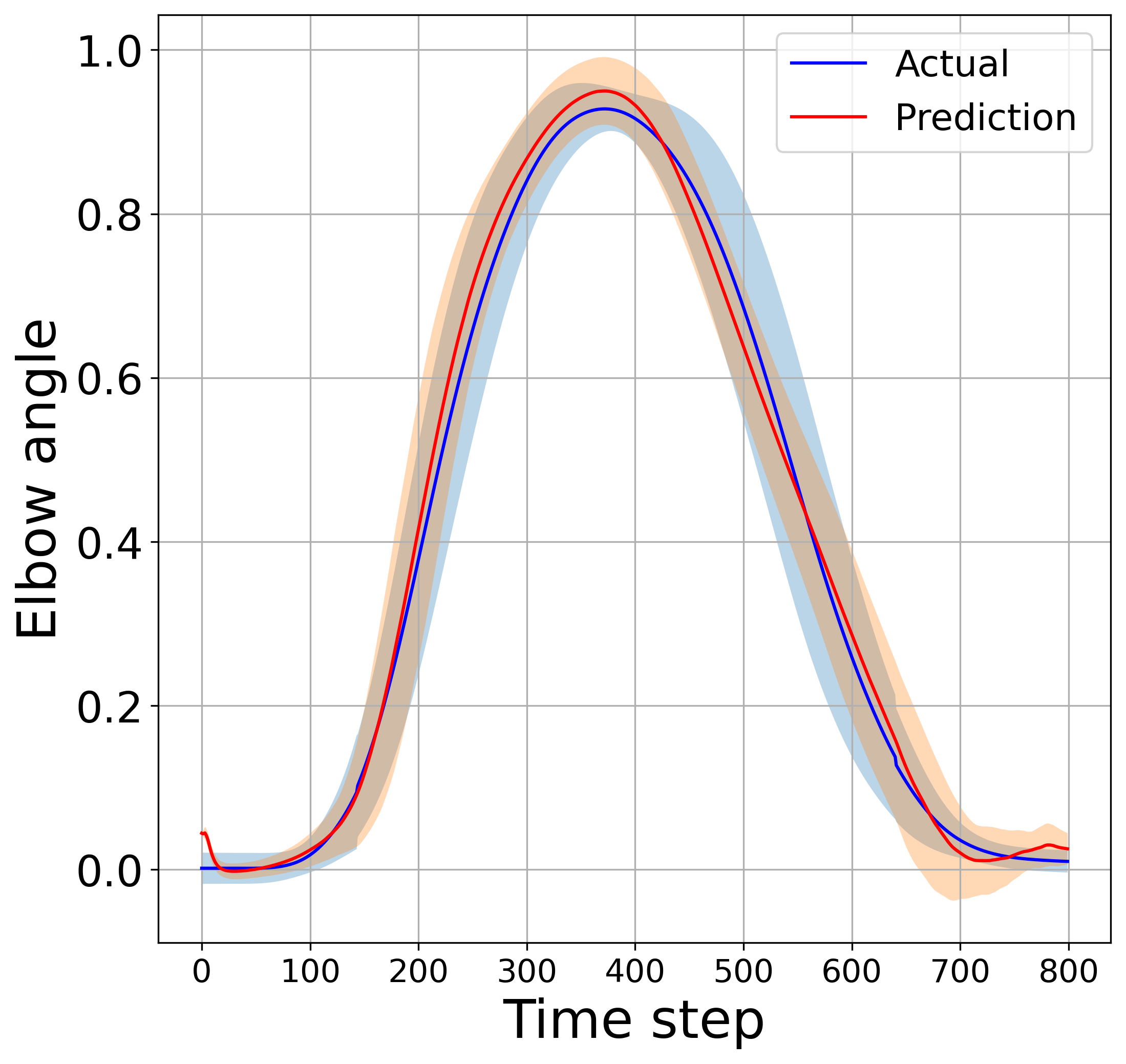}
		\caption{(0.001, 0.046)}
	\end{subfigure}
	\begin{subfigure}{.16\linewidth}
		\includegraphics[width=\linewidth]{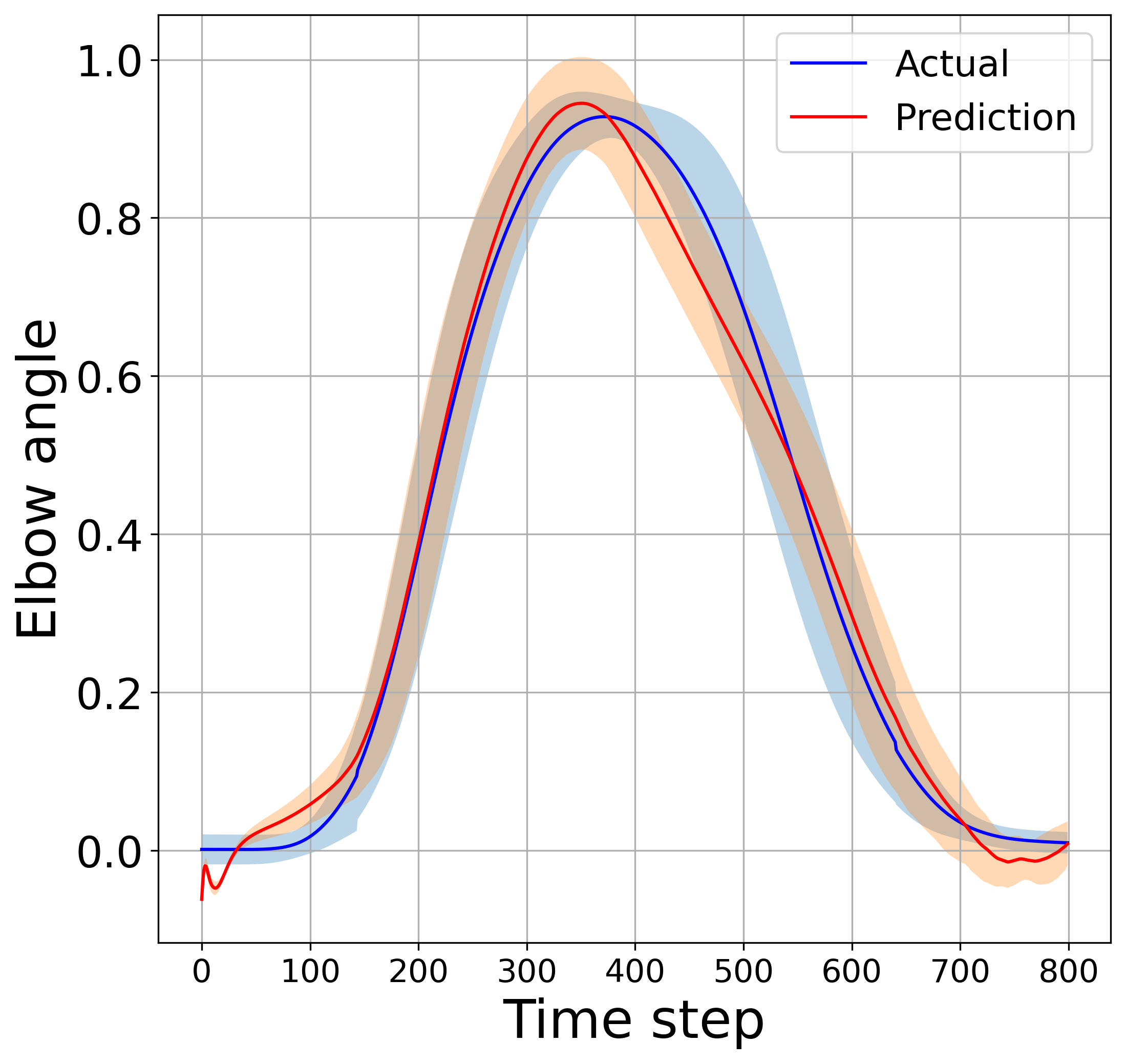}
		\caption{(0.0001, 0.058)}
	\end{subfigure}
	\begin{subfigure}{.16\linewidth}
		\includegraphics[width= \linewidth]{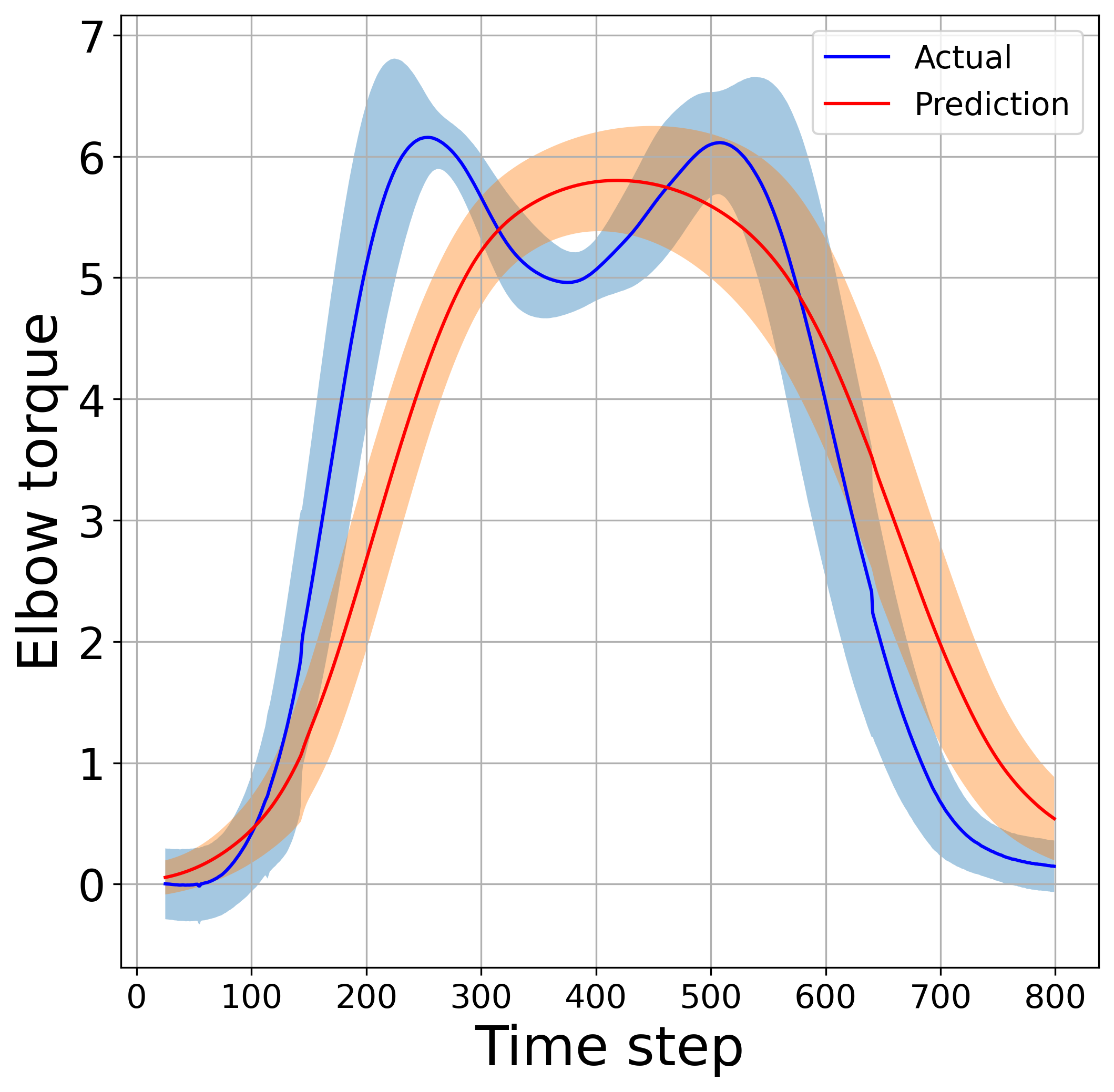}
		\caption{(1.0, 0.16)}
	\end{subfigure}
	\begin{subfigure}{.16\linewidth}
		\includegraphics[width=\linewidth]{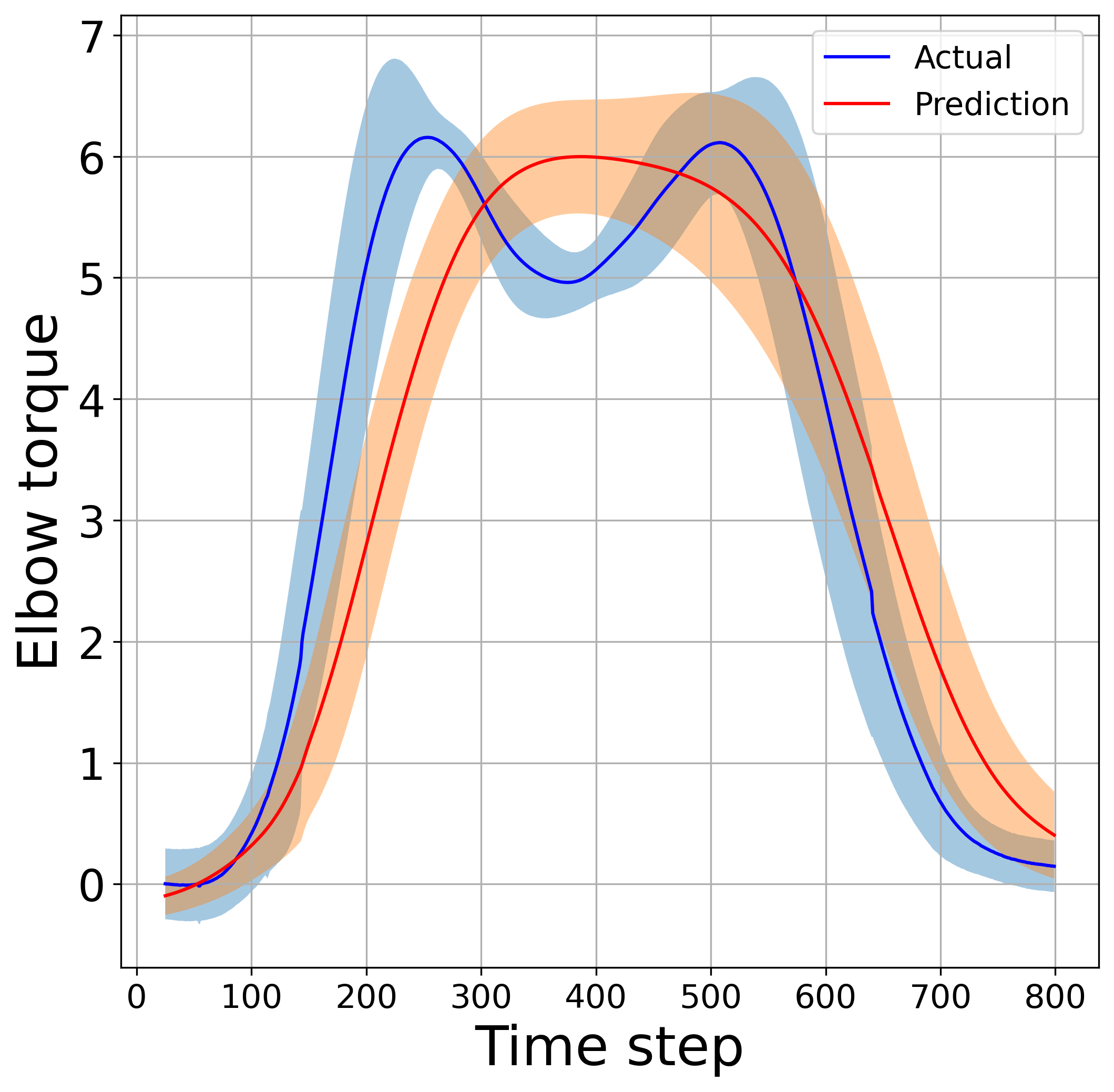}
		\caption{(0.1, 0.15)}
	\end{subfigure}
	\begin{subfigure}{.16\linewidth}
		\includegraphics[width=\linewidth]{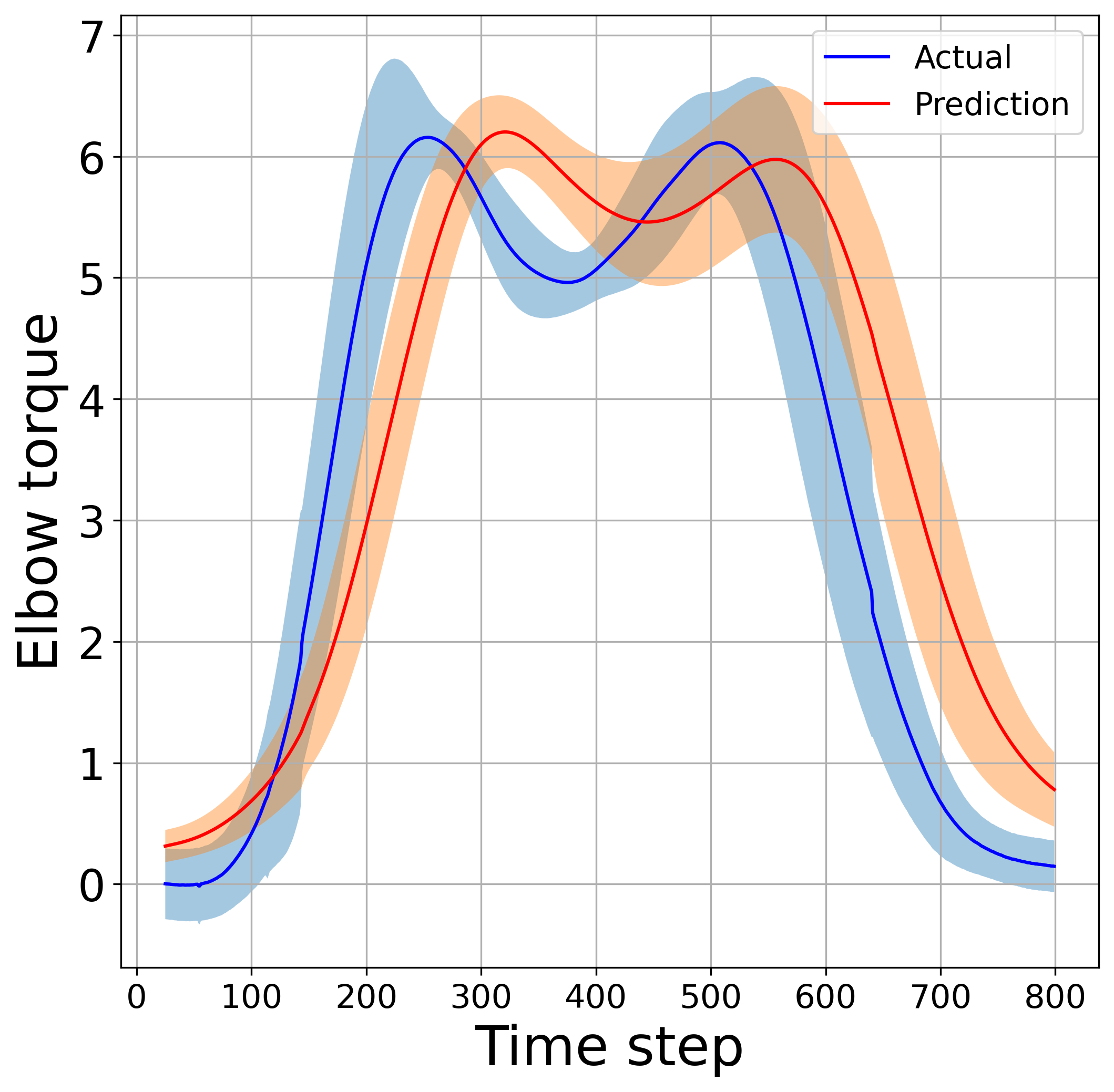}
		\caption{(0.01, 0.14)}
	\end{subfigure}
	\begin{subfigure}{.16\linewidth}
		\includegraphics[width= \linewidth]{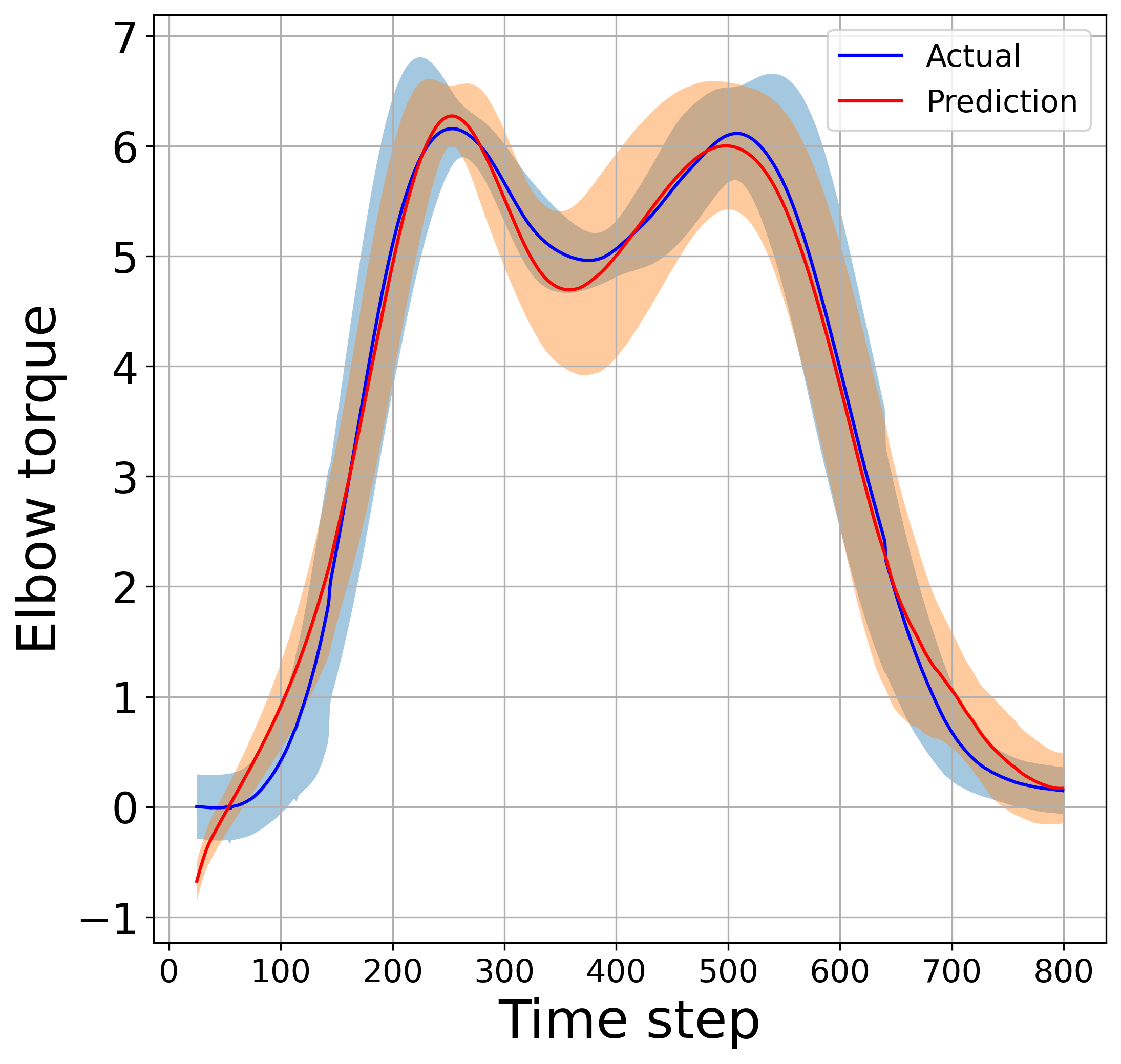}
		\caption{(0.005, 0.096)}
	\end{subfigure}
	\begin{subfigure}{.16\linewidth}
		\includegraphics[width=\linewidth]{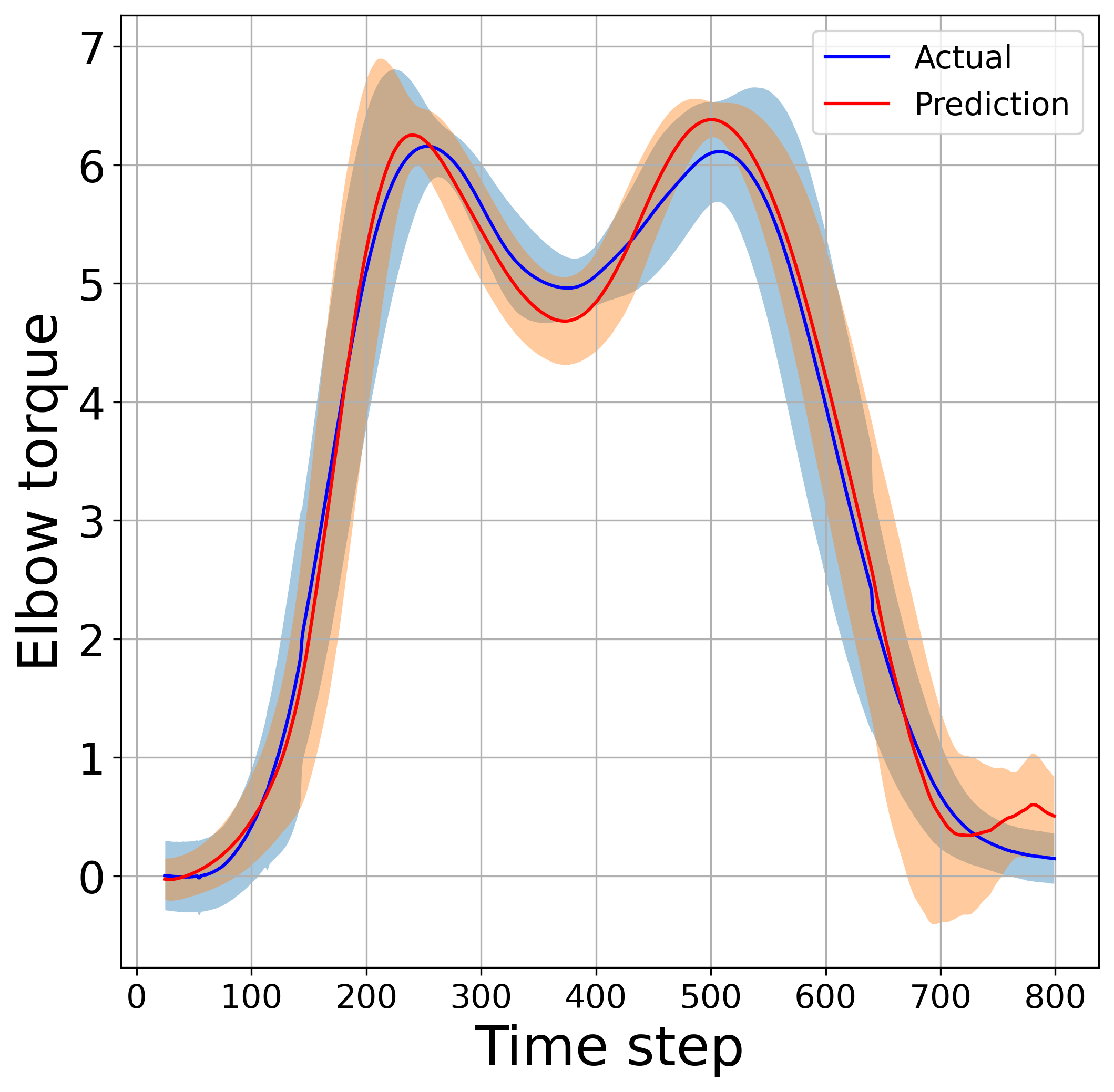}
		\caption{(0.001, 0.075)}
	\end{subfigure}
	\begin{subfigure}{.16\linewidth}
		\includegraphics[width=\linewidth]{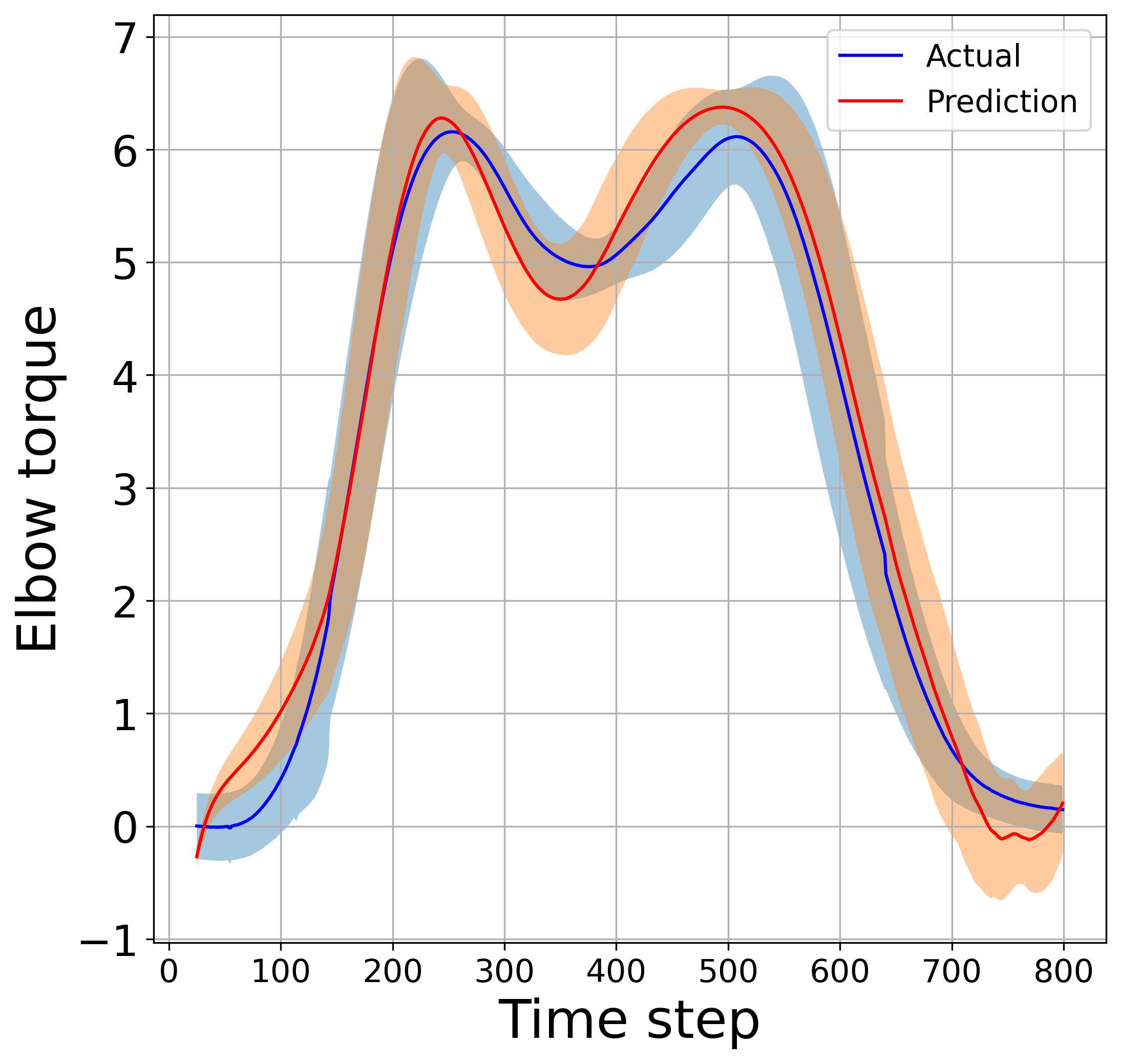}
		\caption{(0.0001, 0.099)}
	\end{subfigure}
	\caption{Representative comparison plots for elbow angle and torque for different values of physics loss weighting factor ($\lambda_{\text{physics}}$). Tuple in caption represents ($\lambda_{\text{physics}}$, RMSE score)}.
    \label{fig:effect of lambda}
\end{figure*}
\begin{figure*}[ht]
	\centering
	\begin{subfigure}{.3\linewidth}
		\includegraphics[width= \linewidth]{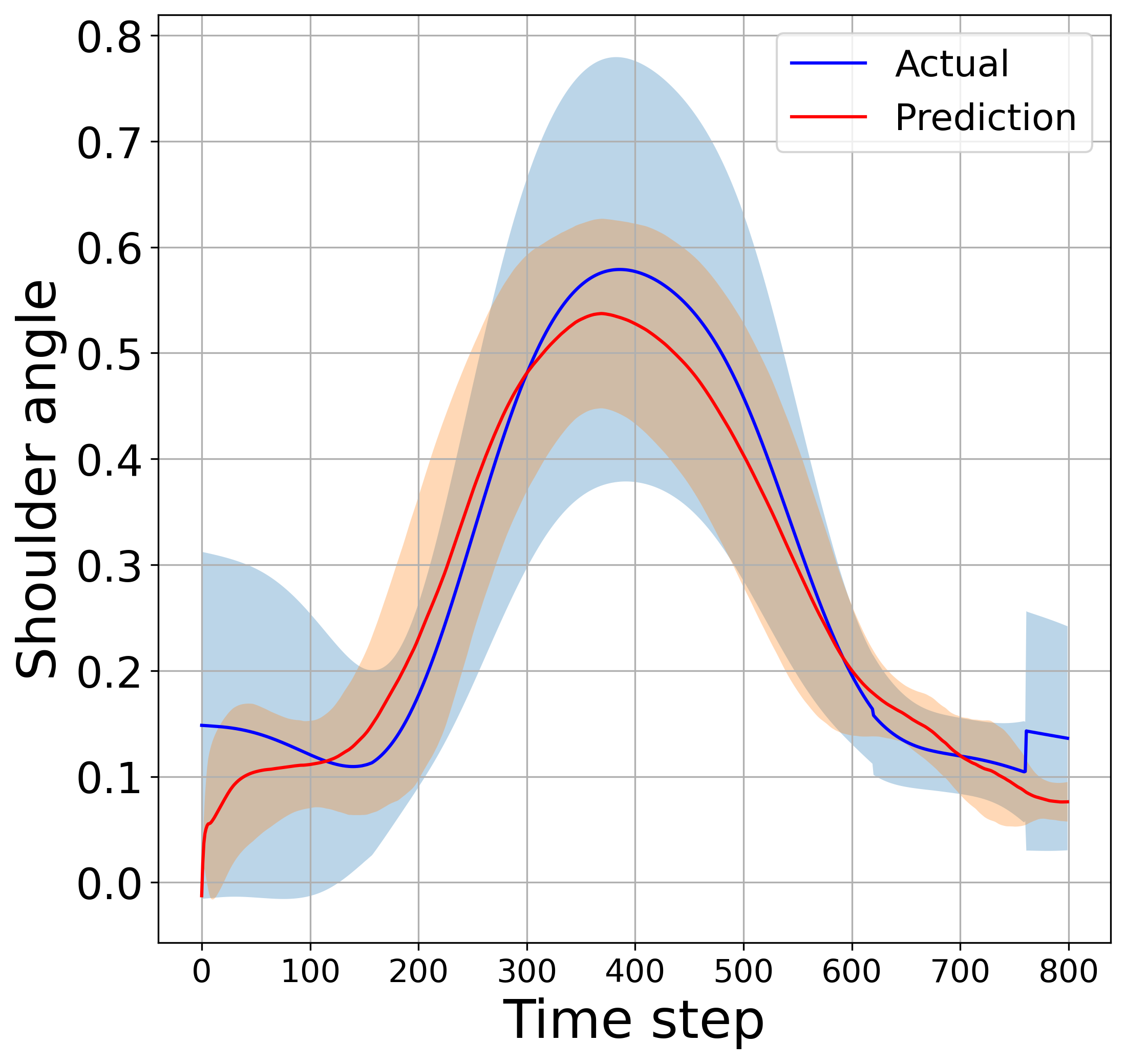}
		\caption{0KG load }
	\end{subfigure}
	\begin{subfigure}{.3\linewidth}
		\includegraphics[width=\linewidth]{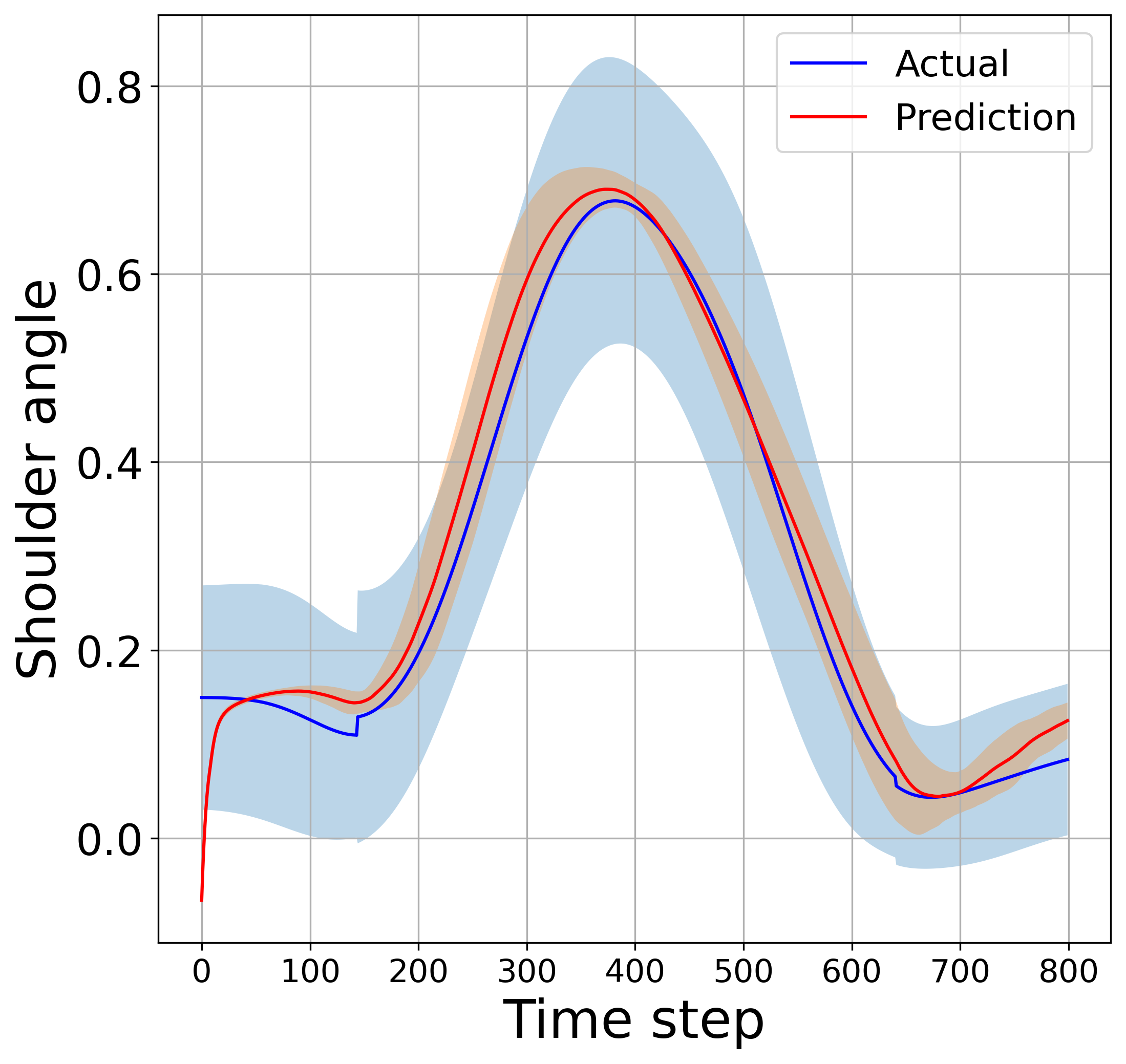}
		\caption{2KG load }
	\end{subfigure}
	\begin{subfigure}{.3\linewidth}
		\includegraphics[width=\linewidth]{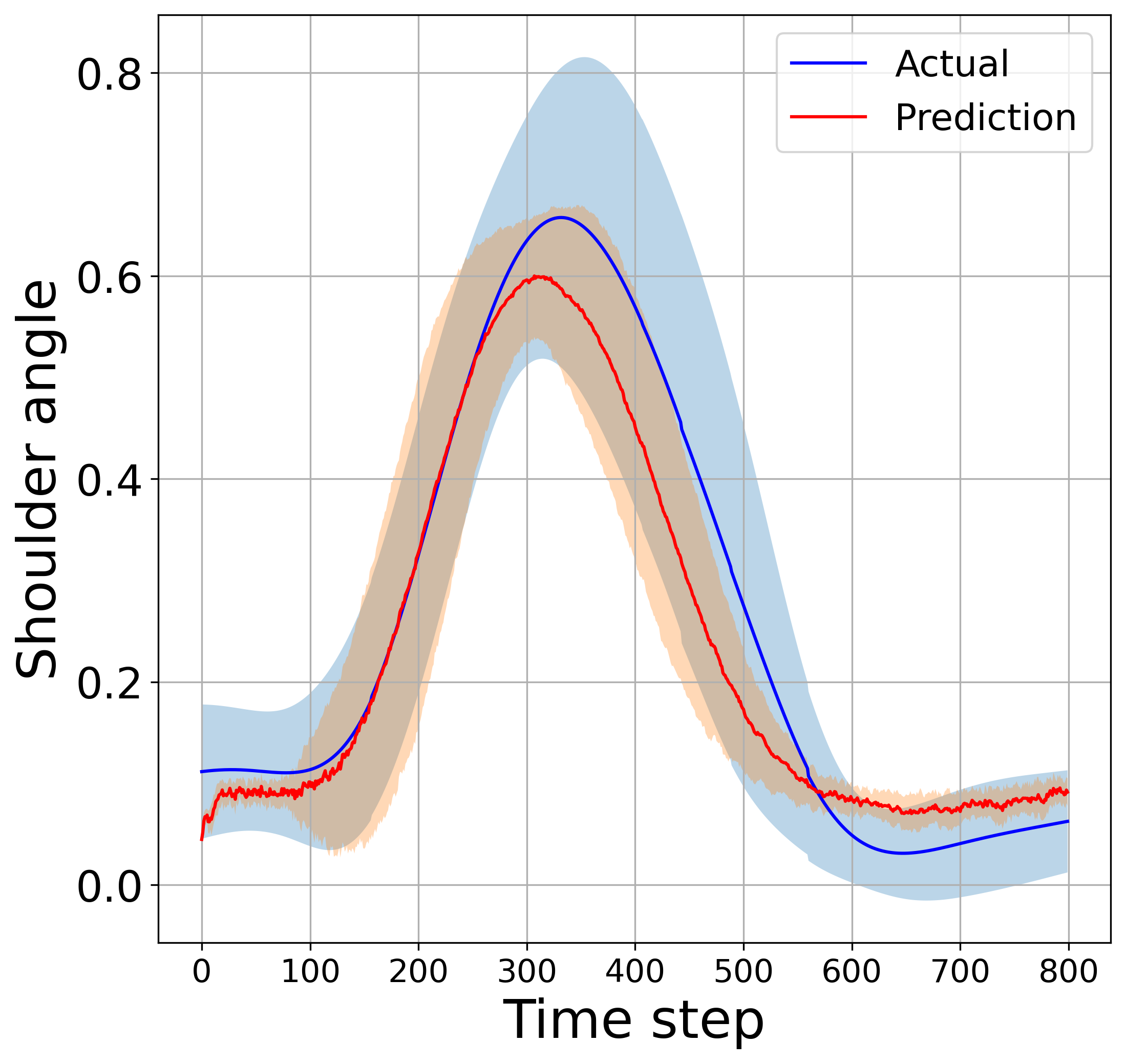}
		\caption{4KG load}
	\end{subfigure}
	\begin{subfigure}{.3\linewidth}
		\includegraphics[width= \linewidth]{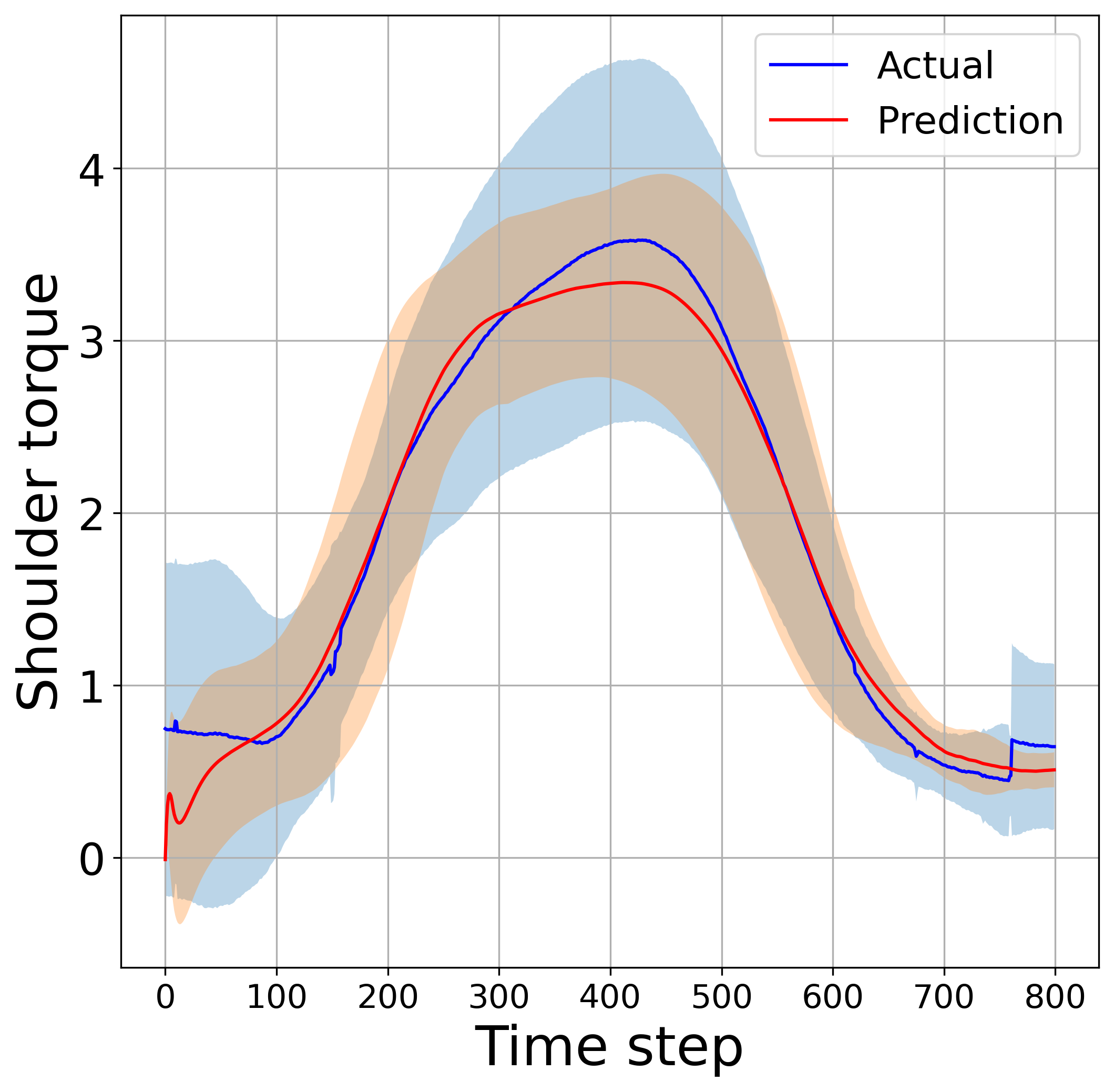}
		\caption{0KG load}
	\end{subfigure}
	\begin{subfigure}{.3\linewidth}
		\includegraphics[width=\linewidth]{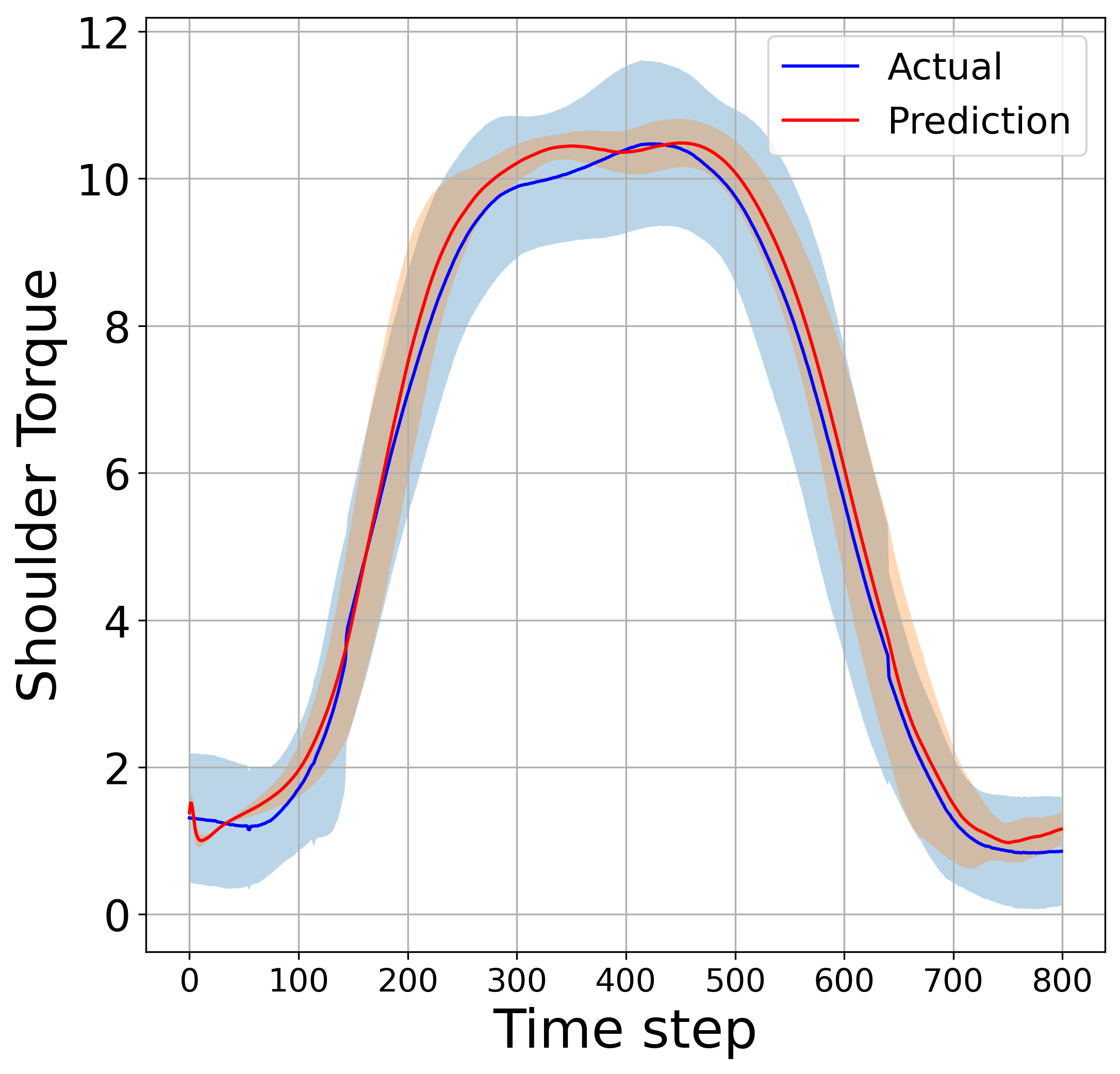}
		\caption{2KG load}
	\end{subfigure}
	\begin{subfigure}{.3\linewidth}
		\includegraphics[width=\linewidth]{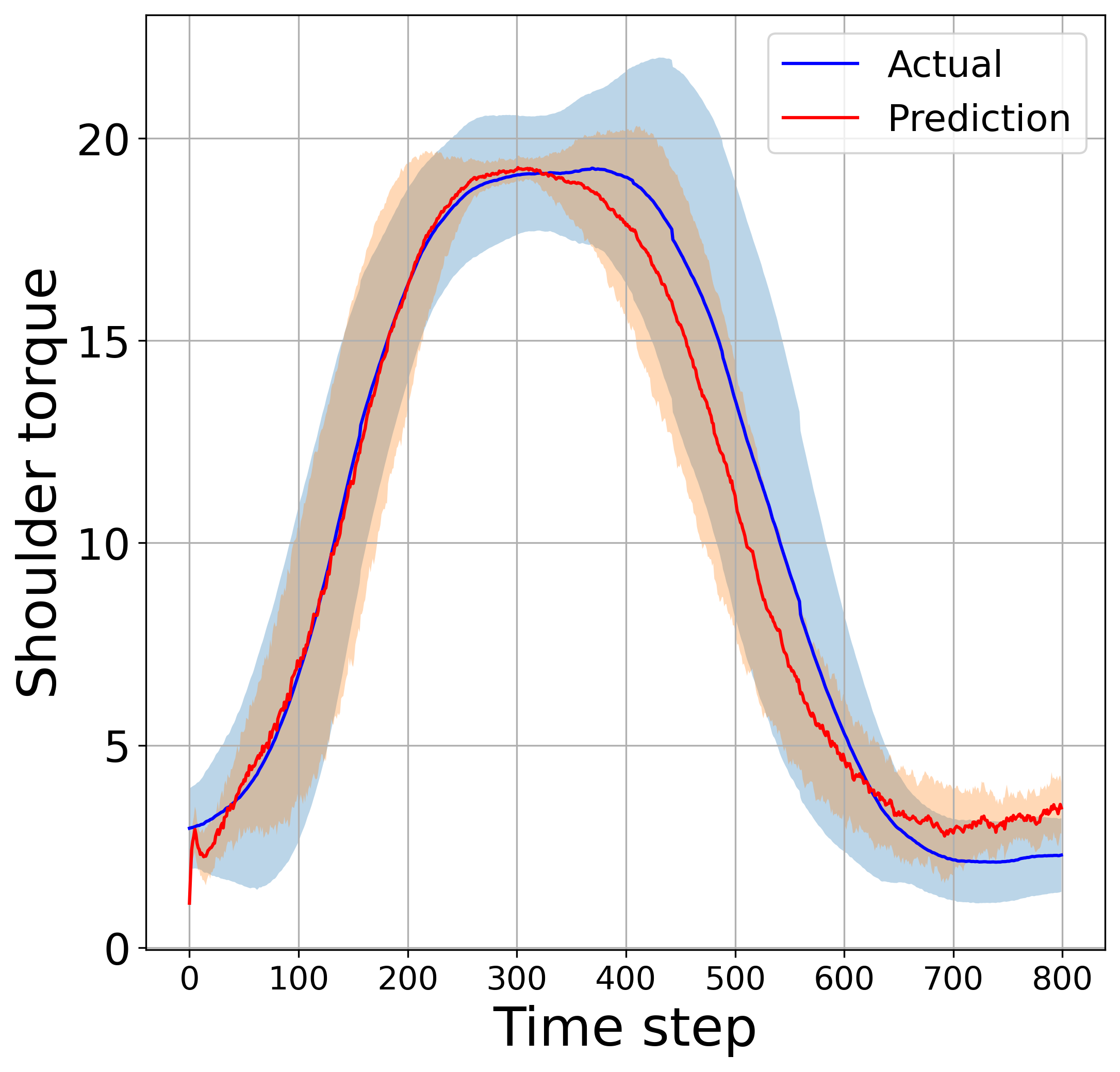}
		\caption{4KG load}
	\end{subfigure}
	\caption{Plots for Shoulder joint angle and Shoulder joint torque prediction of a subject at different loading at hand. Units of shoulder torque is N-m where as shoulder angles are normalized in range of (0-1) whose maximum value is about ${32}^0$.Solid line represents mean value of predicted $\&$ true value of all testing dataset whereas shaded region represents spread of them.}
    \label{fig:Shoulder joint angle and torque prediction}
\end{figure*}
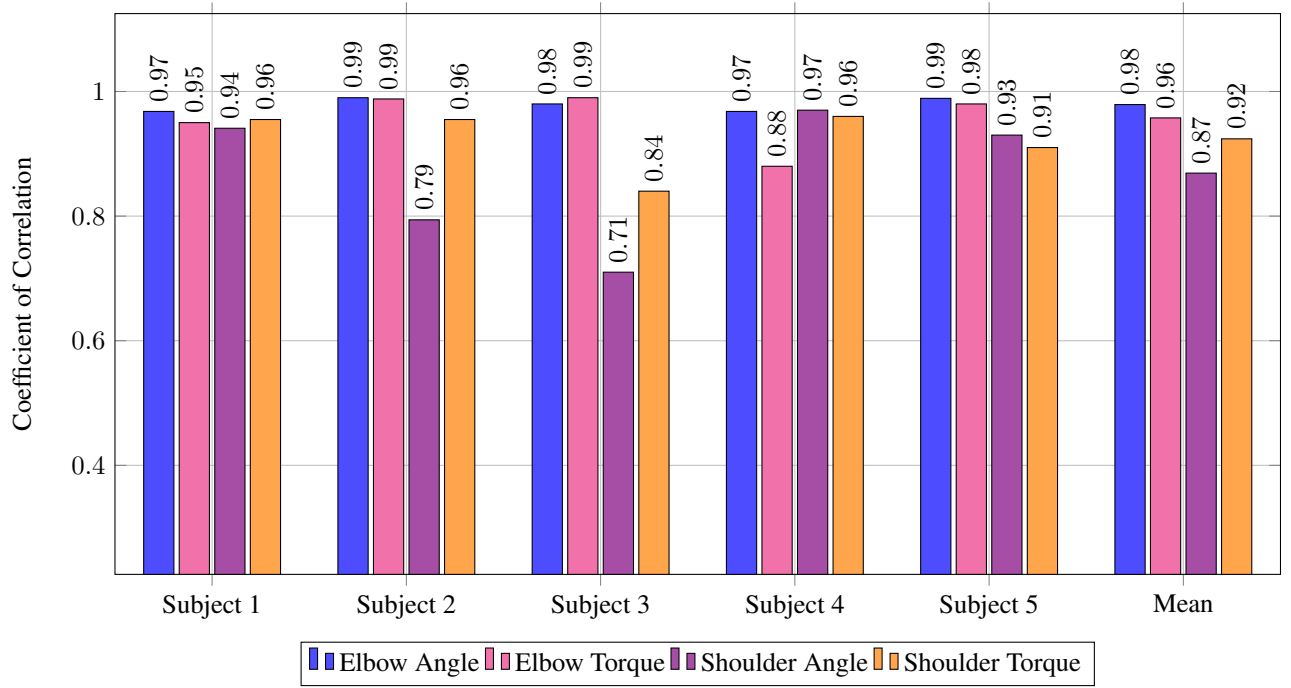
\begin{figure*}[ht]
\centering
\begin{tikzpicture}
\begin{axis}[
    ybar,
    ylabel={Coefficient of Correlation},
    height=9cm, width=17cm,
    enlargelimits=0.1,
    legend style={at={(0.5,-0.12)},
	anchor=north,legend columns=-1},
    ymin=0.30, ymax=1.05,
    bar width = 0.4cm,
    symbolic x coords={Subject 1,Subject 2,Subject 3,Subject 4,Subject 5,Mean},
    xtick=data,
    nodes near coords,
    nodes near coords align={vertical},
    every node near coord/.append style={rotate=90, anchor=west},
    grid = both
    ]
\addplot[fill = blue!70] coordinates {(Subject 1,0.968) (Subject 2,0.99) (Subject 3,0.98 ) (Subject 4,0.968) (Subject 5, 0.989)(Mean,0.979)};
\addplot[fill = magenta!70] coordinates {(Subject 1, 0.95) (Subject 2,.988) (Subject 3,0.99) (Subject 4,0.88) (Subject 5, 0.98)(Mean,0.9576)};
\addplot[fill = violet!70] coordinates {(Subject 1, 0.941) (Subject 2,0.794) (Subject 3,0.71) (Subject 4,0.97) (Subject 5, 0.93)(Mean,0.869)};
\addplot[fill = orange!70] coordinates {(Subject 1, 0.955) (Subject 2,.955) (Subject 3,0.84) (Subject 4,0.96) (Subject 5, 0.91)(Mean,0.924)};

\legend{Elbow Angle,Elbow Torque,Shoulder Angle , Shoulder Torque}
\end{axis}
\end{tikzpicture}
\caption{Coefficient of correlation scores for prediction of Elbow $\&$ Shoulder joint angles $\&$ joint torques for five different subjects using PiGRN model.}
\label{fig:Coeff. of correlation for different subjects}
\end{figure*}
\begin{figure*}[ht!]
	\centering
	\begin{subfigure}{.49\textwidth}
		\includegraphics[clip=true,width= \textwidth]{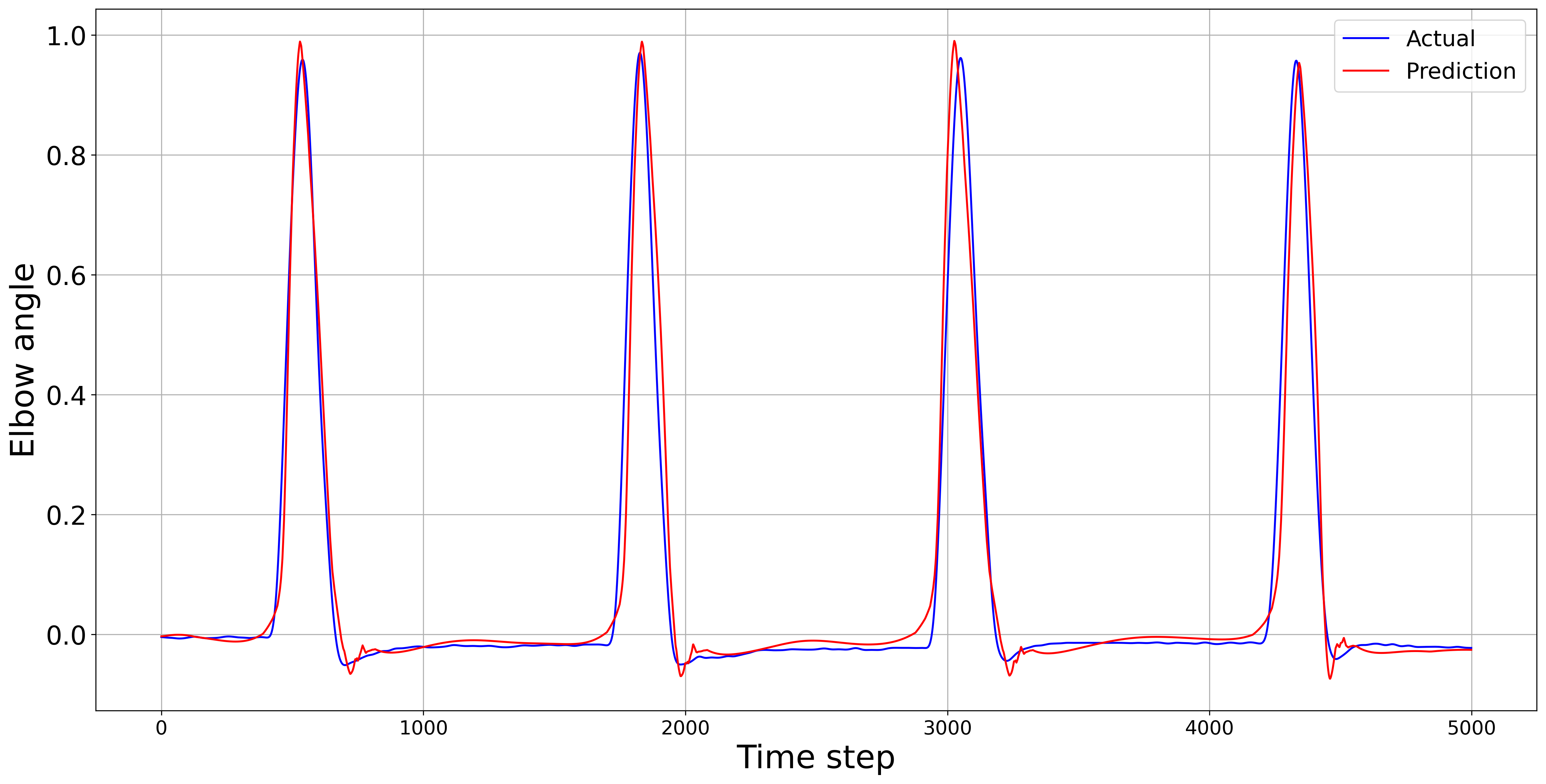}
		\caption{Elbow joint angle }
	\end{subfigure}
	\begin{subfigure}{.49\textwidth}
		\includegraphics[ clip=true,width=\textwidth]{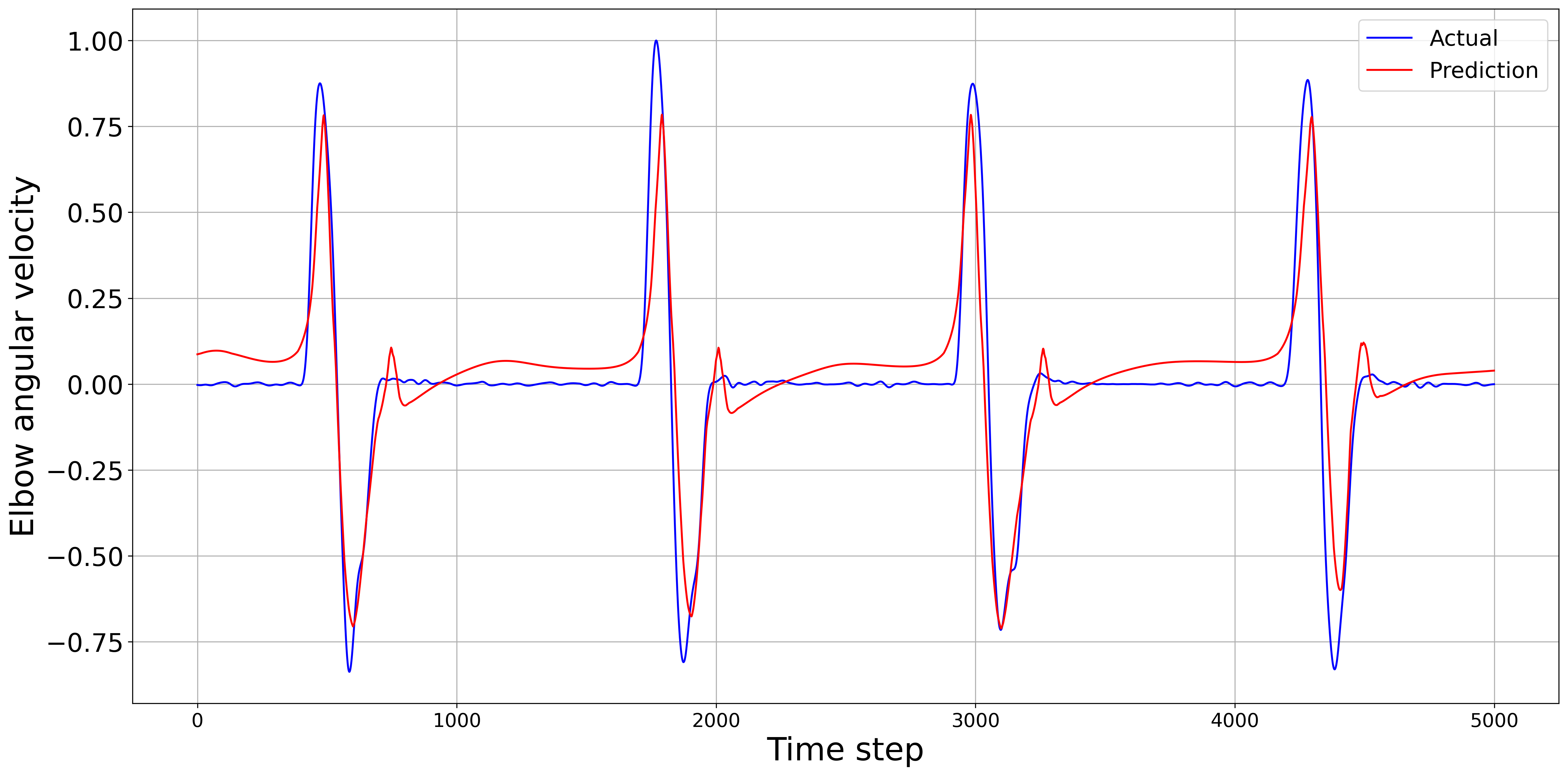}
		\caption{Elbow joint angular velocity }
	\end{subfigure}
	\begin{subfigure}{.49\textwidth}
		\includegraphics[clip=true,width=\textwidth]{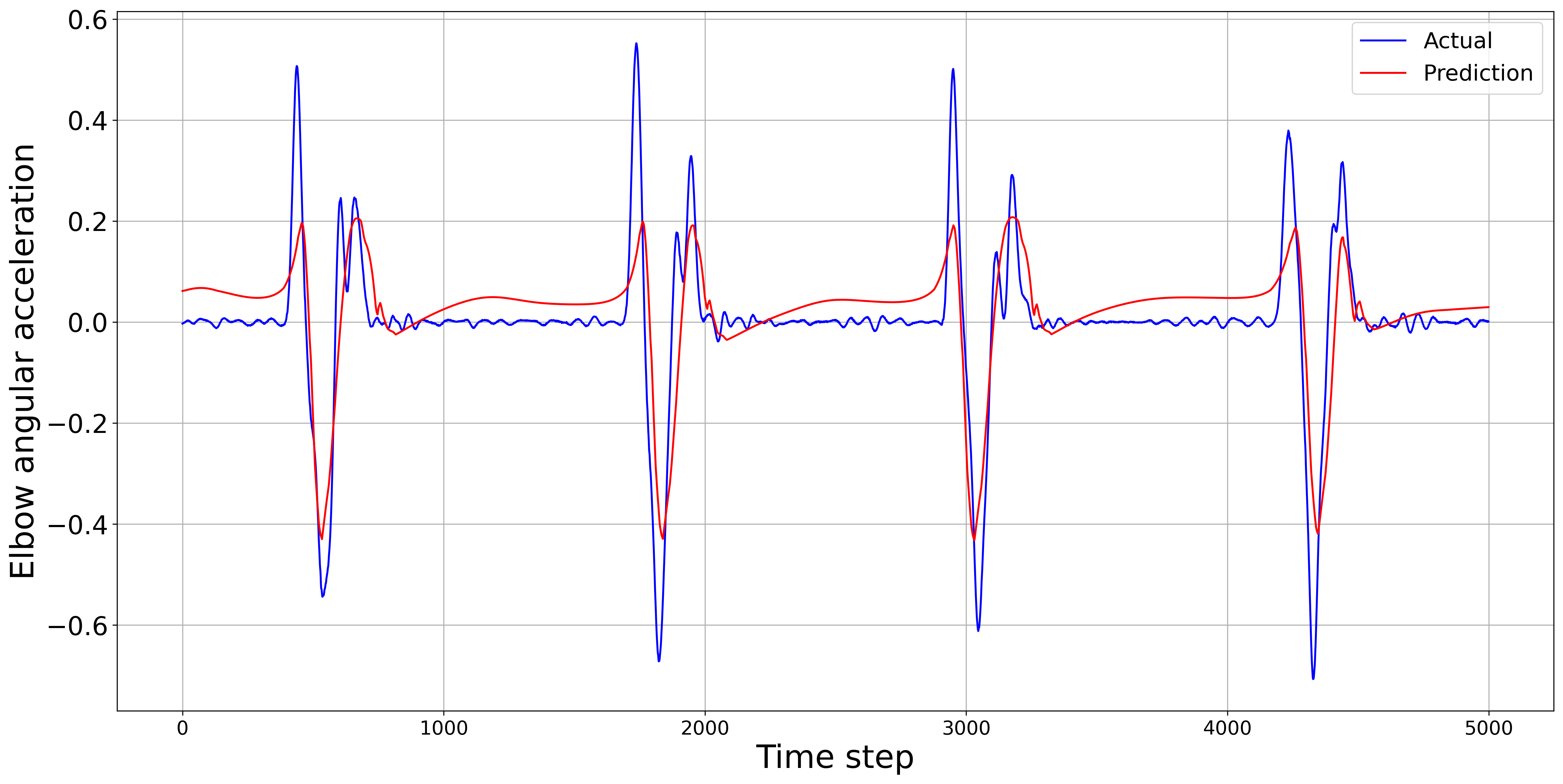}
		\caption{Elbow joint angular acceleration}
	\end{subfigure}
	\begin{subfigure}{.49\textwidth}
		\includegraphics[clip=true,width=\textwidth]{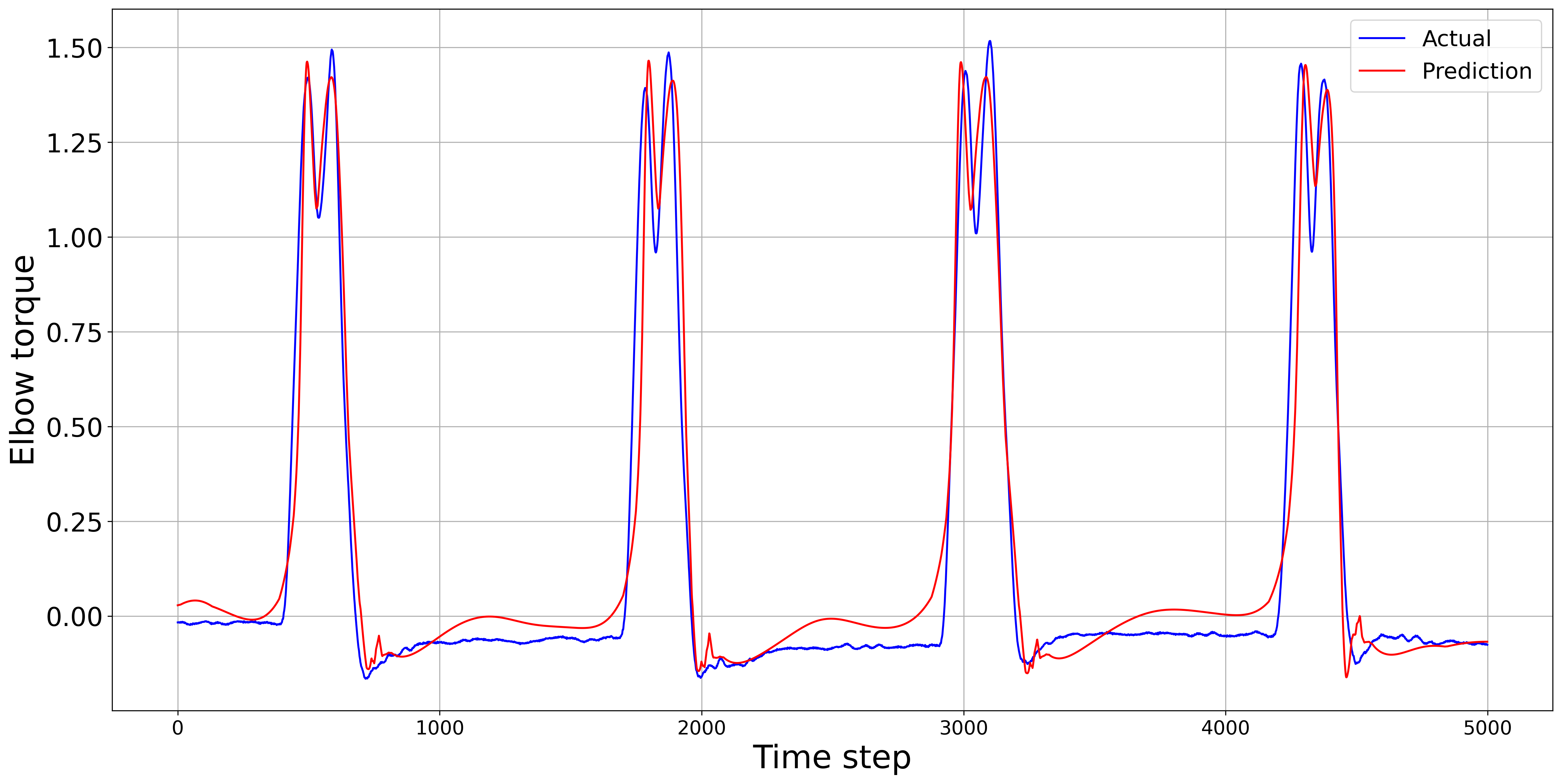}
		\caption{Elbow joint angular acceleration}
	\end{subfigure}
	\caption{Elbow Joint angle, angular velocity, angular acceleration $\&$ torque prediction for long sequence data}
    \label{fig:sample joint parameters}
\end{figure*}
\vspace{-3mm}
\subsection{Evaluation of PiGRN training process}
To illustrate the convergence of the proposed PiGRN framework, we present the convergence of the combined total loss, which encompasses both data loss and physics loss for the elbow, shoulder joint, and external load during training, as illustrated in Fig. \ref{fig:loss curve}. When predicting the joint kinematic parameters for the elbow, shoulder joint, and external load, the total loss significantly decreases after 150 epochs, displaying minor local oscillations as training progresses, and fully converges after 1250 epochs. We attribute this behavior primarily to the batch size being set to 1 during GRU training. This batch size improves the GRU's ability to learn the temporal feature of sEMG data more effectively, while causing some oscillations.

\subsubsection{Overall comparison}
The effectiveness of the proposed method in predicting joint angles and torque using the testing dataset was assessed to verify its robustness. Figures \ref{fig: elbow angle and torque prediction} and \ref{fig:Shoulder joint angle and torque prediction} showcase the outcomes of the proposed framework for elbow and shoulder joints, displaying both joint torques and angles. 
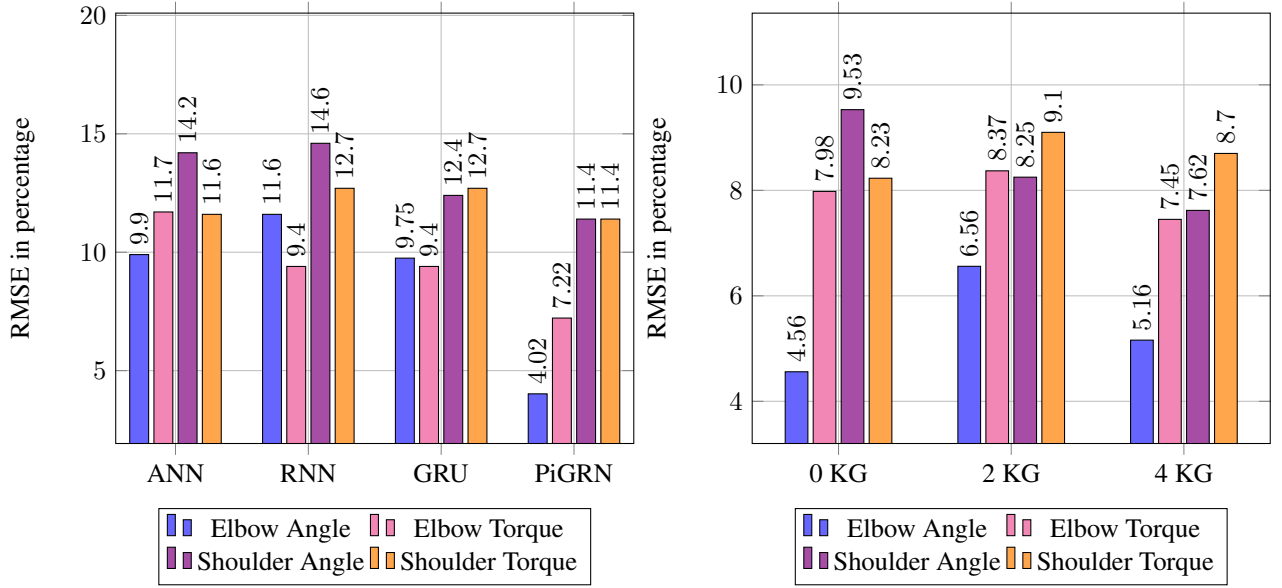
\begin{figure}[ht]
    \centering
    \begin{subfigure}[b]{0.49\textwidth}
        \centering
        \begin{tikzpicture}
        \begin{axis}[
            ybar,
            ylabel = {RMSE in percentage},
            enlargelimits=0.15,
            legend style={at={(0.5, -0.15)}, anchor=north,legend columns=2},
            bar width=0.25cm,
            ymax=18,
            symbolic x coords={ANN, RNN, GRU, PiGRN},
            xtick = data,
            x tick label style={
                /pgf/number format/1000 sep=},
            nodes near coords,
            nodes near coords align={vertical},
            every node near coord/.append style={rotate=90, anchor=west},
            grid = both,
            ]
        \addplot[fill = blue!60] coordinates {(ANN,9.9 ) (RNN, 11.6) (GRU,9.75) (PiGRN, 4.02)};
        \addplot[fill = magenta!60] coordinates {(ANN,11.7) (RNN, 9.4) (GRU,9.4) (PiGRN, 7.22)};
        \addplot[fill = violet!70] coordinates {(ANN,14.2) (RNN, 14.6) (GRU, 12.4) (PiGRN, 11.4)};
        \addplot[fill = orange!70] coordinates {(ANN,11.6) (RNN, 12.7) (GRU, 12.7) (PiGRN, 11.4)};
        \legend{Elbow Angle, Elbow Torque, Shoulder Angle, Shoulder Torque}
        \end{axis}
        \end{tikzpicture}
        \caption{Performance of other models in prediction of joint angle \& torques and their comparison with the current model (PiGRN)}
        \label{fig:Performance of different models}
    \end{subfigure}
    \hfill
    \begin{subfigure}[b]{0.49\textwidth}
        \centering
        \begin{tikzpicture}
        \begin{axis}[
            ybar,
            ylabel={RMSE in percentage},
            enlargelimits=0.25,
            ymax = 10,
            legend style={at={(0.5,-0.15)}, anchor=north,legend columns=2},
            bar width = 0.3cm,
            symbolic x coords={0 KG, 2 KG, 4 KG},
            xtick=data,
            nodes near coords,
            nodes near coords align={vertical},
            every node near coord/.append style={rotate=90, anchor=west},
            grid = both
            ]
        \addplot[fill = blue!60] coordinates {(0 KG, 4.56) (2 KG, 6.56) (4 KG,5.16 )};
        \addplot[fill = magenta!60] coordinates {(0 KG, 7.98) (2 KG, 8.37) (4 KG,7.45 )};
        \addplot[fill = violet!70] coordinates {(0 KG, 9.53) (2 KG, 8.25) (4 KG,7.62)};
        \addplot[fill = orange!70] coordinates {(0 KG, 8.23) (2 KG, 9.1) (4 KG,8.7)};
        \legend{Elbow Angle, Elbow Torque, Shoulder Angle, Shoulder Torque}
        \end{axis}
        \end{tikzpicture}
        \caption{RMSE scores for prediction of joint angle \& torque of a subject across different loadings at hand}
        \label{fig:RMSE scores of Subject1 for different loadings}
    \end{subfigure}
    \caption{Comparison of RMSE scores between different models and different loadings}
    \label{fig:Combined RMSE Comparisons}
\end{figure}
As evidenced in these figures, the predicted values closely match the actual ones, underscoring the remarkable tracking accuracy of the proposed approach. Furthermore, Figure \ref{fig:RMSE scores of five subjects} presents the RMSE values (in \%) across five subjects, while Figure \ref{fig:Coeff. of correlation for different subjects} illustrates the Pearson coefficient of correlation, comparing the predicted elbow angle, elbow torque, shoulder angle, and shoulder torque of the proposed framework with those of baseline methods across different subjects. These figures also presents the mean RMSE values for all five subjects.
\subsubsection{Comparison across different load}
To investigate the performance of the proposed framework across different loading scenarios, we evaluate the proposed framework by constructing a testing dataset consists of 0, 2, 4kg load sEMG data. Figure \ref{fig:RMSE scores of Subject1 for different loadings} demonstrates the detailed result of \% RMSE scores for one subject, we can find that the proposed framework can achieve satisfactory performance.
\subsubsection{Comparison with different methods}
To evaluate the effectiveness of the proposed framework quantitatively, we conduct comprehensive comparisons with baseline methods, as depicted in Figure \ref{fig:Performance of different models}. The findings reveal that our framework consistently yields lower \% RMSEs when forecasting elbow angle, elbow torque, shoulder angle, and shoulder torque, indicating its robustness. 

Notably, when compared to deep learning techniques like GRU, RNN, and ANN, our framework demonstrates superior performance owing to its innate ability to automatically extract sophisticated features from the input data. Among these deep learning approaches, PiGRN emerges as the top performer, which we attribute to the integration of fundamental physics principles that regulate the GRU mechanism in our framework. This incorporation enhances performance beyond traditional MSE loss.
\subsection{Evaluation of external load and joint kinematic parameters prediction}
Figure \ref{fig:sample joint parameters} illustrate the representative results of our proposed framework in predicting joint kinematic parameters. These figures demonstrate the capability of our framework in extracting features related to muscle loading, as well as joint angular velocities and accelerations, from diverse load sEMG data.
\subsection{Evaluation of long sequence joint angle and torque prediction}
To assess the flexibility of our proposed framework in handling varying input sequence lengths and corresponding output sequences, Figure \ref{fig:sample joint parameters} demonstrates long sequence predictions of elbow joint angle and joint torque. These results indicate that our trained model is capable of processing input sequences of different lengths beyond those it was initially trained on.
\subsection{Effect of physics informed weighting factor ($\lambda_{\text{physics}}$)}
We conducted an investigation into the impact of the physics loss weighting factor ($\lambda_{\text{physics}}$) on the performance of our proposed framework, focusing on scenarios involving elbow joint angle and torque. The detailed results are presented in Figure \ref{fig:effect of lambda}. Specifically, we examined six physics-informed weighting factors: 1.0, 0.1, 0.01, 0.05, 0.001, and 0.0001, while keeping other hyperparameters constant. Analysis of Figure \ref{fig:effect of lambda} reveals that the proposed framework achieves optimal performance with a smaller weighting factor of 0.001.
\section{DISCUSSION AND FUTURE WORK}
In this research, we employed a sophisticated physics-informed gated recurrent network (PiGRN), combining a GRU time-series architecture with the differential equation of motion for the upper limb. This hybrid model predicts joint kinematic parameters at the elbow and shoulder joints and the external load, which are then used to predict joint torque. Unlike ANN or CNN models, our framework allows flexible input and output, making it suitable for real-time joint torque prediction in HMI development. During training, we used 800 timesteps of data to speed up the process. However, once trained, the model can handle input sequences of any length. Figure \ref{fig:sample joint parameters} shows the prediction of joint elbow angle and torque for 5000 timesteps, demonstrating the PiGRN model's generalization capability. Currently, we predict the elbow and shoulder joints with two degrees of freedom (DOF). Future work will expand to more joints and DOF, potentially covering the entire upper or lower limb.

For the physics-informed component, we utilized the equations of motion, where the joint angles, angular velocities, and angular accelerations serve as inputs. These inputs are predicted from the GRU component itself.  This approach reduces the time needed for automatic differentiation as well. Figure \ref{fig:sample joint parameters} showcases a prediction plot for elbow angle, angular velocity, and elbow angular acceleration as an illustrative example. From the figure, it is evident that the predictions closely match the actual values, exhibiting sharp patterns. Thus, the proposed PiGRN model successfully captures the intricate patterns of joint kinematic parameters once trained, incorporating physical laws into its architecture and training process. This ensures that the network's predictions align with established physical principles. It is worth noting that in this study, actual values for the joint kinematic parameters were computed using the central difference method. However, in future applications, joint angular accelerations can be measured experimentally using sensors such as inertial measurement units (IMUs), from which angular velocities can be computed.

During our framework implementation, we found that predicting shoulder angle and torque isn't as accurate as elbow angle and torque, resulting in higher RMSE scores (Fig. \ref{fig:RMSE scores of five subjects}). Our focus was on predicting multi-joint dynamics in sagittal plane movement only, where shoulder angle was smaller than elbow angle during the biceps curl-like motion. For this, we used sEMG from the biceps and triceps, neglecting deltoid muscle involvement. Future work could include deltoid muscle sEMG to improve shoulder prediction accuracy.

\section{CONCLUSION}
This paper presents a novel physics-informed gated recurrent neural network (PiGRN) developed to predict multi-joint dynamics using sEMG data under various load conditions. The proposed method is capable of predicting multi-joint kinematic parameters and external loads from sEMG sequences of different lengths. Unlike conventional deep learning approaches, our framework uniquely integrates predicted joint kinematic parameters and external loads into physics-based domain knowledge as soft constraints to regularize the GRU's loss function, a technique that has not been implemented before. This strategy reduces the dependence on automatic differentiation and decreases training time. Experiments with sEMG data across three different load groups demonstrate the method's effectiveness in predicting joint torque and kinematics. We believe this framework could serve as a general method for predicting joint torque and kinematics, making it well-suited for real-time applications in the development of robust HMIs. Although this study focused on data involving two degrees of freedom, two joints, and in-plane movements, future work will aim to address more generalized datasets and musculoskeletal models, as well as explore unsupervised learning techniques.
\section*{Acknowledgment}
The authors would like to thank Anant Jain from Indian Institute of Technology (IIT) Delhi for helping in data collection. This study is partly supported by the Joint Advanced Technology Centre (JATC) (Grant no.: RP03830G) and the I-Hub Foundation for cobotics (IHFC) section-8 company (Grant no.: GP/2021/RR/010) at IIT Delhi, sponsored by the Ministry of Education (MoE), Govt. of India.


\begin{thebibliography}{10}

\bibitem{xia2024shaping}
Haisheng Xia, Yuchong Zhang, Nona Rajabi, Farzaneh Taleb, Qunting Yang, Danica Kragic, and Zhijun Li.
\newblock Shaping high-performance wearable robots for human motor and sensory reconstruction and enhancement.
\newblock {\em Nature Communications}, 15(1):1--12, 2024.

\bibitem{kapsalyamov2020state}
Akim Kapsalyamov, Shahid Hussain, and Prashant~K Jamwal.
\newblock State-of-the-art assistive powered upper limb exoskeletons for elderly.
\newblock {\em IEEE Access}, 8:178991--179001, 2020.

\bibitem{zhang2021development}
Fuhai Zhang, Lei Yang, and Yili Fu.
\newblock Development and test of a spasm sensor for hand rehabilitation exoskeleton.
\newblock {\em IEEE Transactions on Instrumentation and Measurement}, 71:1--8, 2021.

\bibitem{esposito2021biosignal}
Daniele Esposito, Jessica Centracchio, Emilio Andreozzi, Gaetano~D Gargiulo, Ganesh~R Naik, and Paolo Bifulco.
\newblock Biosignal-based human--machine interfaces for assistance and rehabilitation: A survey.
\newblock {\em Sensors}, 21(20):6863, 2021.

\bibitem{kwon2011real}
Suncheol Kwon and Jung Kim.
\newblock Real-time upper limb motion estimation from surface electromyography and joint angular velocities using an artificial neural network for human--machine cooperation.
\newblock {\em IEEE transactions on Information Technology in Biomedicine}, 15(4):522--530, 2011.

\bibitem{young2016state}
D~Young and D~Ferris.
\newblock State-of-the-art and future direction for robotic lower limb exoskeletons.
\newblock {\em IEEE Trans. On Neural System and Rehabilitation Engineering}, 2016.

\bibitem{tucker2015control}
Michael~R Tucker, Jeremy Olivier, Anna Pagel, Hannes Bleuler, Mohamed Bouri, Olivier Lambercy, Jos{\'e} del~R Mill{\'a}n, Robert Riener, Heike Vallery, and Roger Gassert.
\newblock Control strategies for active lower extremity prosthetics and orthotics: a review.
\newblock {\em Journal of neuroengineering and rehabilitation}, 12:1--30, 2015.

\bibitem{zheng2022surface}
Mingde Zheng, Michael~S Crouch, and Michael~S Eggleston.
\newblock Surface electromyography as a natural human--machine interface: a review.
\newblock {\em IEEE Sensors Journal}, 22(10):9198--9214, 2022.

\bibitem{durandau2019voluntary}
Guillaume Durandau, Dario Farina, Guillermo As{\'\i}n-Prieto, Iris Dimbwadyo-Terrer, Sergio Lerma-Lara, Jose~L Pons, Juan~C Moreno, and Massimo Sartori.
\newblock Voluntary control of wearable robotic exoskeletons by patients with paresis via neuromechanical modeling.
\newblock {\em Journal of neuroengineering and rehabilitation}, 16:1--18, 2019.

\bibitem{pizzolato2015ceinms}
Claudio Pizzolato, David~G Lloyd, Massimo Sartori, Elena Ceseracciu, Thor~F Besier, Benjamin~J Fregly, and Monica Reggiani.
\newblock Ceinms: A toolbox to investigate the influence of different neural control solutions on the prediction of muscle excitation and joint moments during dynamic motor tasks.
\newblock {\em Journal of biomechanics}, 48(14):3929--3936, 2015.

\bibitem{lloyd2003emg}
David~G Lloyd and Thor~F Besier.
\newblock An emg-driven musculoskeletal model to estimate muscle forces and knee joint moments in vivo.
\newblock {\em Journal of biomechanics}, 36(6):765--776, 2003.

\bibitem{cheron1996dynamic}
Guy Cheron, J-P Draye, Mark Bourgeios, and Ga{\"e}tan Libert.
\newblock A dynamic neural network identification of electromyography and arm trajectory relationship during complex movements.
\newblock {\em IEEE Transactions on Biomedical Engineering}, 43(5):552--558, 1996.

\bibitem{liu1999dynamic}
Ming~Ming Liu, Walter Herzog, and Hans~HCM Savelberg.
\newblock Dynamic muscle force predictions from emg: an artificial neural network approach.
\newblock {\em Journal of electromyography and kinesiology}, 9(6):391--400, 1999.

\bibitem{au2000emg}
Arthur~TC Au and Robert~F Kirsch.
\newblock Emg-based prediction of shoulder and elbow kinematics in able-bodied and spinal cord injured individuals.
\newblock {\em IEEE Transactions on rehabilitation engineering}, 8(4):471--480, 2000.

\bibitem{subasi2006classification}
Abdulhamit Subasi, Mustafa Yilmaz, and Hasan~Riza Ozcalik.
\newblock Classification of emg signals using wavelet neural network.
\newblock {\em Journal of neuroscience methods}, 156(1-2):360--367, 2006.

\bibitem{hahn2007feasibility}
Michael~E Hahn.
\newblock Feasibility of estimating isokinetic knee torque using a neural network model.
\newblock {\em Journal of Biomechanics}, 40(5):1107--1114, 2007.

\bibitem{song2005using}
Rong Song and Kai-Yu Tong.
\newblock Using recurrent artificial neural network model to estimate voluntary elbow torque in dynamic situations.
\newblock {\em Medical and Biological Engineering and Computing}, 43:473--480, 2005.

\bibitem{chandrapal2011investigating}
Mervin Chandrapal, XiaoQi Chen, WenHui Wang, Benjamin Stanke, and Nicolas Le~Pape.
\newblock Investigating improvements to neural network based emg to joint torque estimation.
\newblock {\em Paladyn, Journal of Behavioral Robotics}, 2(4):185--192, 2011.

\bibitem{aung2013estimation}
Yee~Mon Aung and Adel Al-Jumaily.
\newblock Estimation of upper limb joint angle using surface emg signal.
\newblock {\em International Journal of Advanced Robotic Systems}, 10(10):369, 2013.

\bibitem{ngeo2014continuous}
Jimson~G Ngeo, Tomoya Tamei, and Tomohiro Shibata.
\newblock Continuous and simultaneous estimation of finger kinematics using inputs from an emg-to-muscle activation model.
\newblock {\em Journal of neuroengineering and rehabilitation}, 11:1--14, 2014.

\bibitem{xia2018emg}
Peng Xia, Jie Hu, and Yinghong Peng.
\newblock Emg-based estimation of limb movement using deep learning with recurrent convolutional neural networks.
\newblock {\em Artificial organs}, 42(5):E67--E77, 2018.

\bibitem{chen2019continuous}
Yan Chen, Song Yu, Ke~Ma, Shuangyuan Huang, Guofeng Li, Siqi Cai, and Longhan Xie.
\newblock A continuous estimation model of upper limb joint angles by using surface electromyography and deep learning method.
\newblock {\em IEEE Access}, 7:174940--174950, 2019.

\bibitem{huang2020joint}
Yanjiang Huang, Kaibin Chen, Xianmin Zhang, Kai Wang, and Jun Ota.
\newblock Joint torque estimation for the human arm from semg using backpropagation neural networks and autoencoders.
\newblock {\em Biomedical Signal Processing and Control}, 62:102051, 2020.

\bibitem{zhang2022lower}
Longbin Zhang, Davit Soselia, Ruoli Wang, and Elena~M Gutierrez-Farewik.
\newblock Lower-limb joint torque prediction using lstm neural networks and transfer learning.
\newblock {\em IEEE Transactions on Neural Systems and Rehabilitation Engineering}, 30:600--609, 2022.

\bibitem{solares2020deep}
Jose Roberto~Ayala Solares, Francesca Elisa~Diletta Raimondi, Yajie Zhu, Fatemeh Rahimian, Dexter Canoy, Jenny Tran, Ana Catarina~Pinho Gomes, Amir~H Payberah, Mariagrazia Zottoli, Milad Nazarzadeh, et~al.
\newblock Deep learning for electronic health records: A comparative review of multiple deep neural architectures.
\newblock {\em Journal of biomedical informatics}, 101:103337, 2020.

\bibitem{kim2019subject}
Keun-Tae Kim, Cuntai Guan, and Seong-Whan Lee.
\newblock A subject-transfer framework based on single-trial emg analysis using convolutional neural networks.
\newblock {\em IEEE Transactions on Neural Systems and Rehabilitation Engineering}, 28(1):94--103, 2019.

\bibitem{zhang2022boosting}
Jie Zhang, Yihui Zhao, Tianzhe Bao, Zhenhong Li, Kun Qian, Alejandro~F Frangi, Sheng~Quan Xie, and Zhi-Qiang Zhang.
\newblock Boosting personalized musculoskeletal modeling with physics-informed knowledge transfer.
\newblock {\em IEEE Transactions on Instrumentation and Measurement}, 72:1--11, 2022.

\bibitem{karniadakis2021physics}
George~Em Karniadakis, Ioannis~G Kevrekidis, Lu~Lu, Paris Perdikaris, Sifan Wang, and Liu Yang.
\newblock Physics-informed machine learning.
\newblock {\em Nature Reviews Physics}, 3(6):422--440, 2021.

\bibitem{meng2022physics}
Chuizheng Meng, Sungyong Seo, Defu Cao, Sam Griesemer, and Yan Liu.
\newblock When physics meets machine learning: A survey of physics-informed machine learning.
\newblock {\em arXiv preprint arXiv:2203.16797}, 2022.

\bibitem{zhang2022physics}
Jie Zhang, Yihui Zhao, Fergus Shone, Zhenhong Li, Alejandro~F Frangi, Sheng~Quan Xie, and Zhi-Qiang Zhang.
\newblock Physics-informed deep learning for musculoskeletal modeling: Predicting muscle forces and joint kinematics from surface emg.
\newblock {\em IEEE Transactions on Neural Systems and Rehabilitation Engineering}, 31:484--493, 2022.

\bibitem{zhang2023towards}
Jie Zhang, Ziling Ruan, Qing Li, and Zhi-Qiang Zhang.
\newblock Towards robust and efficient musculoskeletal modelling using distributed physics-informed deep learning.
\newblock {\em IEEE Transactions on Instrumentation and Measurement}, 2023.

\bibitem{taneja2022feature}
Karan Taneja, Xiaolong He, QiZhi He, Xinlun Zhao, Yun-An Lin, Kenneth~J Loh, and Jiun-Shyan Chen.
\newblock A feature-encoded physics-informed parameter identification neural network for musculoskeletal systems.
\newblock {\em Journal of biomechanical engineering}, 144(12):121006, 2022.

\bibitem{shi2023physics}
Yue Shi, Shuhao Ma, Yihui Zhao, Chaoyang Shi, and Zhiqiang Zhang.
\newblock A physics-informed low-shot adversarial learning for semg-based estimation of muscle force and joint kinematics.
\newblock {\em IEEE Journal of Biomedical and Health Informatics}, 2023.

\bibitem{ma2024physics}
Shuhao Ma, Jie Zhang, Chaoyang Shi, Pei Di, Ian~D Robertson, and Zhi-Qiang Zhang.
\newblock Physics-informed deep learning for muscle force prediction with unlabeled semg signals.
\newblock {\em IEEE Transactions on Neural Systems and Rehabilitation Engineering}, 2024.

\bibitem{delp2007opensim}
Scott~L Delp, Frank~C Anderson, Allison~S Arnold, Peter Loan, Ayman Habib, Chand~T John, Eran Guendelman, and Darryl~G Thelen.
\newblock Opensim: open-source software to create and analyze dynamic simulations of movement.
\newblock {\em IEEE transactions on biomedical engineering}, 54(11):1940--1950, 2007.

\bibitem{cho2014learning}
Kyunghyun Cho, Bart Van~Merri{\"e}nboer, Caglar Gulcehre, Dzmitry Bahdanau, Fethi Bougares, Holger Schwenk, and Yoshua Bengio.
\newblock Learning phrase representations using rnn encoder-decoder for statistical machine translation.
\newblock {\em arXiv preprint arXiv:1406.1078}, 2014.

\bibitem{chung2014empirical}
Junyoung Chung, Caglar Gulcehre, KyungHyun Cho, and Yoshua Bengio.
\newblock Empirical evaluation of gated recurrent neural networks on sequence modeling.
\newblock {\em arXiv preprint arXiv:1412.3555}, 2014.

\bibitem{noh2021analysis}
Seol-Hyun Noh.
\newblock Analysis of gradient vanishing of rnns and performance comparison.
\newblock {\em Information}, 12(11):442, 2021.

\bibitem{nosouhian2021review}
Shiva Nosouhian, Fereshteh Nosouhian, and Abbas~Kazemi Khoshouei.
\newblock A review of recurrent neural network architecture for sequence learning: Comparison between lstm and gru.
\newblock 2021.

\bibitem{shewalkar2019performance}
Apeksha Shewalkar, Deepika Nyavanandi, and Simone~A Ludwig.
\newblock Performance evaluation of deep neural networks applied to speech recognition: Rnn, lstm and gru.
\newblock {\em Journal of Artificial Intelligence and Soft Computing Research}, 9(4):235--245, 2019.

\bibitem{herman2016physics}
Irving~P Herman.
\newblock {\em Physics of the human body}.
\newblock Springer, 2016.

\bibitem{hinrichs1985regression}
Richard~N Hinrichs.
\newblock Regression equations to predict segmental moments of inertia from anthropometric measurements: an extension of the data of chandler et al.(1975).
\newblock {\em Journal of Biomechanics}, 18(8):621--624, 1985.

\bibitem{konrad2005abc}
Peter Konrad.
\newblock The abc of emg.
\newblock {\em A practical introduction to kinesiological electromyography}, 1(2005):30--5, 2005.

\bibitem{shrirao2009neural}
Nikhil~A Shrirao, Narender~P Reddy, and Durga~R Kosuri.
\newblock Neural network committees for finger joint angle estimation from surface emg signals.
\newblock {\em Biomedical engineering online}, 8:1--11, 2009.

\bibitem{tang2015impact}
Zhichuan Tang, Hongnian Yu, and Shuang Cang.
\newblock Impact of load variation on joint angle estimation from surface emg signals.
\newblock {\em IEEE Transactions on Neural Systems and Rehabilitation Engineering}, 24(12):1342--1350, 2015.

\end{thebibliography}

\end{document}